\documentclass{article}

\usepackage{microtype}
\usepackage{graphicx}
\usepackage{subfigure}
\usepackage{booktabs}
\usepackage{hyperref}
\usepackage{mathrsfs}
\usepackage{amsmath}
\usepackage{amssymb}
\usepackage{mathtools}
\usepackage{amsthm}
\usepackage[capitalize,noabbrev]{cleveref}
\usepackage{dsfont}
\usepackage{amsmath}
\usepackage{xcolor,lipsum,subcaption}
\usepackage{booktabs}   
\usepackage{xcolor}     
\usepackage{siunitx}     
\usepackage{amssymb}
\usepackage{tikz}
\usepackage{circuitikz}
\usepackage{xcolor}

\usepackage[accepted]{icml2025}

\definecolor{mylightcoral}{RGB}{245, 116, 112}
\definecolor{myblue}{RGB}{96, 165, 199}
\definecolor{mybrown}{RGB}{178,80,77}
\definecolor{mylightyellow}{RGB}{233,233,217}
\definecolor{mygreen}{RGB}{149,237,148}
\definecolor{mypurple}{RGB}{128,0,128}

\theoremstyle{plain}
\newtheorem{theorem}{Theorem}[section]
\newtheorem{proposition}[theorem]{Proposition}

\theoremstyle{definition}

\newtheorem{assumption}[theorem]{Assumption}
\theoremstyle{remark}

\usepackage[textsize=tiny]{todonotes}

\newcommand{\xPast}{x_{1:t_0-1}} 
\newcommand{\xFuture}{x_{t_0:T}} 
\newcommand{\xTotal}{x_{1:T}}
\newcommand{\XPast}{\mathcal{X}^{1:t_0-1}} 
\newcommand{\XFuture}{\mathcal{X}^{t_0:T}} 
\newcommand{\dimn}{D} 
\newcommand{\tT}{t_0:T}
\newcommand{\tPast}{1:t_{0}-1}
\newcommand{\ctT}{c_{1:T}}

\newcommand{\Yk}{\mathcal{X}_k}
\newcommand{\Lk}{\mathcal{L}^{k}_{\theta}}
\newcommand{\Ls}{\mathcal{L}^{s}_{\theta}}
\newcommand{\Lr}{\mathcal{L}^{r}_{\theta}}

\icmltitlerunning{Winner-takes-all for Multivariate Probabilistic Time Series Forecasting}

\begin{document}

\twocolumn[
\icmltitle{Winner-takes-all for Multivariate Probabilistic Time Series Forecasting}
\icmlsetsymbol{equal}{*}

\begin{icmlauthorlist}
\icmlauthor{Adrien Cortés}{equal}
\icmlauthor{Rémi Rehm}{}
\icmlauthor{Victor Letzelter}{equal,telecom,valeo}
\end{icmlauthorlist}

\icmlaffiliation{telecom}{LTCI, Télécom Paris, Institut Polytechnique de Paris, France}
\icmlaffiliation{valeo}{Valeo.ai, Paris, France}

\icmlcorrespondingauthor{Adrien Cortés}{ad.cortes43@gmail.com}
\icmlcorrespondingauthor{Victor Letzelter}{letzelter.victor@hotmail.fr}

\icmlkeywords{Time-series forecasting, Probabilistic methods, Conditional Distribution Estimation, Winner-takes-all}

\vskip 0.3in
]
\printAffiliationsAndNotice{\icmlEqualContribution}

\begin{abstract}
We introduce \texttt{TimeMCL}, a method leveraging the Multiple Choice Learning (MCL) paradigm to forecast multiple plausible time series futures. Our approach employs a neural network with multiple heads and utilizes the Winner-Takes-All (WTA) loss to promote diversity among predictions. MCL has recently gained attention due to its simplicity and ability to address ill-posed and ambiguous tasks. We propose an adaptation of this framework for time-series forecasting, presenting it as an efficient method to predict diverse futures, which we relate to its implicit \textit{quantization} objective. We provide insights into our approach using synthetic data and evaluate it on real-world time series, demonstrating its promising performance at a light computational cost.
\end{abstract}

\section{Introduction}

Predicting the weather of the upcoming weekend or the stock prices of next month with perfect accuracy would undoubtedly be useful. Unfortunately, time-series forecasting is a highly ill-posed problem. In many cases, the available input information is insufficient to reduce our uncertainty about the estimation of the underlying stochastic process and the data itself may contain noise. Consequently, the best a forecaster can do is estimate \emph{plausible} future trajectories, ideally along with the probability of each outcome.

Because temporal data are highly structured and typically come with input–output pairs that require no additional manual annotation, autoregressive neural networks have become the de facto standard for forecasting high-dimensional time series from historical data and exogenous covariates \cite{rangapuram2018deep,salinas2019high,benidis2020neural}. To capture predictive uncertainty, practitioners often place an explicit distribution on the model’s output and perform maximum likelihood estimation \cite{salinas2020deepar, alexandrov2020gluonts}. While such parametric methods can be computationally efficient, they may depend heavily on the choice of distribution family, reducing their flexibility in capturing complex uncertainties \cite{gneiting2014probabilistic}.

In parallel, the success of general-purpose generative models, especially conditional diffusion models \cite{ho2020denoising} such as \texttt{TimeGrad} \cite{rasul2021autoregressive}, has led to strong empirical performance in high-dimensional time series forecasting. However, these models tend to be computationally expensive at inference, particularly when multiple \emph{what-if} scenarios, which we will refer to as \textit{hypotheses}, need to be generated in real-time.  Moreover, there is often no explicit mechanism to guarantee sufficiently diverse hypotheses within a single model pass.\\
\\
To address these limitations, we propose \texttt{TimeMCL}—a novel approach based on Multiple Choice Learning (MCL) techniques—that produces diverse and plausible predictions via a single forward pass.

\paragraph{Contributions.}
\begin{itemize}
    \item We introduce \texttt{TimeMCL}, a new approach for time series forecasting that adapts the Winner-Takes-All (WTA) loss to generate multiple plausible futures.
    \item We show that \texttt{TimeMCL} can be viewed as a \textit{functional quantizer}, and we illustrate its theoretical properties on synthetic data.
    \item We evaluate our method on real-world benchmarks, demonstrating that \texttt{TimeMCL} efficiently produces a diverse set of forecasts with just a few samples in a single forward pass.\footnote{Code available at \href{https://github.com/Victorletzelter/timeMCL}{https://github.com/Victorletzelter/timeMCL}.}
\end{itemize}

\section{Related Work}

\textbf{Ambiguity and need for diversity time series forecasting.}  In recent years, sequence-to-sequence neural networks \cite{hochreiter1997long, seqtoseq14, chung2014empirical, torres2021deep} have become increasingly effective in time series forecasting, often surpassing classical techniques \cite{hyndman2008forecasting,hyndman2008automatic}. Yet capturing the inherent ambiguity in future outcomes remains a critical challenge, especially in high-dimensional settings \cite{ashok2024tactis}. \citet{salinas2020deepar} proposed a probabilistic autoregressive global model capable of fitting and forecasting high-dimensional time series while highlighting the need to quantify uncertainty in predictions. Building on this line of research, \citet{rasul2021autoregressive} introduced a conditional diffusion model that summarizes the past values of the time series into a hidden state, then performs a diffusion process—conditioned on this state—to generate forecasts. \citet{rasul2021multivariate} retained the conditional architecture but replaced the diffusion mechanism with a normalizing-flow generator. While these methods are capable of modeling uncertainty, computational efficiency remains a crucial factor, particularly in real-time scenarios \cite{chen2018real} where multiple plausible futures must be generated. To address these challenges, we introduce a new family of general-purpose autoregressive time series forecasters based on the Winner-Takes-All (WTA) principle, leveraging its quantization properties to produce diverse and realistic forecasts in a single forward.

\textbf{Optimal vector quantization \& Multiple choice learning for conditional distribution estimation.} Quantization is concerned with finding the best finitely supported approximation of a probability measure \cite{bennett1948spectra, du1999centroidal, pages2015introduction, chevallier2018uniform}. In the context of time series forecasting, this translates to quantizing the \emph{conditional} probability distribution of the target time series. Multiple Choice Learning (MCL) with a Winner-Takes-All (WTA) loss \cite{guzman2012multiple, lee2016stochastic} provides a practical framework for such conditional quantization through multi-head networks, which act as a fixed set of codevectors (also called \textit{hypotheses}) \cite{rupprecht2017learning, letzelter24winner, perera2024annealed}. While MCL has thus far been explored in various applications, notably computer vision tasks \cite{lee2017confident, rupprecht2017learning, tian2019versatile}, we adapt it here to predict a quantized representation of the conditional probability distribution of future time series values, using a training scheme specifically tailored for this setting.

\section{Problem setup and notations}
\label{sec:setup}
Let $(x_{t:T}) \in \mathcal{X}^{T-t+1}$ represent a multivariate time series on $\mathcal{X} = \mathbb{R}^{\dimn}$ over time indexes $[\![t,T]\!]$, where $t \in [\![1,T]\!]$. We aim to learn the conditional law 
    \begin{equation}
    p(\xFuture \mid \xPast, \ctT),
    \end{equation}
of future values $\xFuture$ (over the interval \([\![t_{0}, T]\!]\)) given past observations \(\xPast\) and covariates \(\ctT\), the latter being omitted in the following for conciseness. Our focus is on scenarios where the conditional distribution may exhibit multiple modes (\emph{multi-modality}), motivating a richer representation than a single-mean regressor.

To address this issue, the goal of probabilistic time series forecasting is to capture \emph{conditional} distributions over future time series given past values with model $p_{\theta}$, with parameters $\theta$, whose likelihood can be expressed as
\begin{equation}
p_{\theta}(\xFuture \mid \xPast) = \prod_{t = t_0}^{T} p_{\theta}(x_{{t}} \mid x_{{1:t-1}})\;.
\label{eq:conditional_dist}
\end{equation}
Once trained, ancestral sampling methods can be used to infer sequence-level predictions. Let us illustrate this scheme using for instance hidden-variable based recurrent neural networks (RNNs) from \citet{graves2013generating}.

When considering hidden-variables-based models, the basic building block of sequence-to-sequence architectures \cite{seqtoseq14}, one often implicitly parametrize the model with, up to a (log) normalization constant, 
\begin{equation}\mathrm{log}\;p_{\theta}(x_{{t}} \mid x_{{t_{0}:t-1}}, \xPast) = - \ell(f_{\theta}(h_{t-1}),x_t)\;,
\end{equation} where $\ell(\cdot,\cdot)$ be an appropriate loss, \textit{e.g.,} the mean squared error $\ell(\alpha,\beta) \triangleq \| \alpha - \beta \|^{2}$ and all the context and previous states is encapsulated into a hidden state $h_{t-1} \in \mathcal{H}$. Assuming the vanilla form for recurrent networks, the hidden state propagation is represented by a model $s_{\theta}$ with $h_{t-1} = s_{\theta}(x_{t-1},h_{t-2})$, and $f_{\theta}$ is the final projection.

Once trained, the predictions can then be performed by first encoding the past sequence $\xPast$ with a hidden state by applying recursively $s_{\theta}$: $h_{t_{0}} = s_{\theta}(x_{t_{0}-1},\dots, s_{\theta}(x_{2},s_{\theta}(x_{1},h_{0})\dots))$, where $h_{0}$ is an arbitrary initial hidden state. Then the recurrent model can be \textit{unrolled}, \textit{i.e.,} turned into autoregressive mode by decoding the predicted sequence by applying recursively $s_{\theta}$, this time over its own predicted outputs, leading the prediction $\hat{x}_{{t_0}:T} \sim p_{\theta}(x_{{t_0}:T} \mid x_{{t_{0}:T-1}}, h_{t_{0}})$. In the following, we will denote $\mathscr{F}_{\theta}: \xPast \mapsto \hat{x}_{{t_0}:T} = \mathscr{F}_{\theta}(\xPast) \in \mathcal{X}^{T-t_{0}+1}$ the unrolled network.

\section{\texttt{TimeMCL} method}

\texttt{TimeMCL} leverages the \emph{Winner-Takes-All} principle \cite{guzman2012multiple, lee2016stochastic}, which was originally introduced to address ambiguous tasks. WTA naturally extends to scenarios in which future time-series trajectories exhibit \emph{multimodality} (\textit{e.g.,} seasonality, regime switches, and abrupt events). 
We build on the \emph{Multiple Choice Learning} framework, which produces $K$ distinct ``hypotheses'' through multiple heads. Not only does our estimator allow us to produce plausible hypotheses; \texttt{TimeMCL} provably \textit{quantizes} the target distribution of futures, and is therefore expected to infer the $K$ most representative predictions of the target distribution.

\subsection{Training scheme}
\label{sec:WTA-time-series}

A key insight behind WTA is that learning $K$ separate hypotheses 
$\hat{x}^{1}, \dots, \hat{x}^{K} \in \mathcal{X}^{T-t_{0}+1}$ with an objective that effectively induces a \emph{tesselation} \cite{du1999centroidal} of the target space into $K$ cells (one for each hypothesis) aims to capture the best possible information of the target distribution with $K$ points.

\texttt{TimeMCL} works as an alternative to the vanilla maximum likelihood estimation (MLE) of \eqref{eq:conditional_dist}. Let $p_{\theta}^1, \dots, p_{\theta}^{K}$ be $K$ models with parameters $(\theta_1, \dots, \theta_K)$, for which one can associate \textit{heads} $f_{\theta}^{1},\dots,f_{\theta}^{K}: \mathcal{H} \rightarrow \mathcal{X}$ using the hidden-state representation as in Section \ref{sec:setup}. In our implementation, the $K$ models have shared $s_{\theta}$ for hidden-state propagation, and differ only by their final heads $f_{\theta}^k$, and one may define the complete models with $f_{\theta}^k \circ s_{\theta}$.

The Winner-Takes-All consists of the following training scheme for each data point $(\xPast, \xFuture)$: 
\begin{enumerate}
    \item We compute the negative-log-likelihood of each model 
    $$\Lk(\xPast, \xFuture) = - \sum_{t=t_{0}}^{T} \mathrm{log} \; p_{\theta}^{k}(x_{{t}} \mid x_{{1:t-1}})\;,$$
    where 
    $\mathrm{log} \; p_{\theta}^{k}(x_{{t}} \mid x_{{1:t-1}}) = - \ell(f_{\theta}^k(h_{t-1}), x_t)$, for each head $k \in [\![1,K]\!]$.
    \item We pick the ``winner'' $k^{\star} = \arg\min_k \Lk$, and we backpropagate \emph{only} through that winning head ($k^{\star}$).
\end{enumerate}

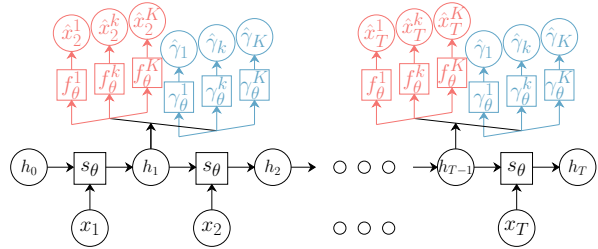
\begin{figure}[h]
    \centering
    \resizebox{0.95\linewidth}{!}{
\begin{circuitikz}
\tikzstyle{every node}=[font=\Huge]
\draw  (6.25,16.5) circle (0.75cm) node {\Huge $x_{1}$} ;
\draw  (11.25,16.5) circle (0.75cm) node {\Huge $x_{2}$} ;
\draw  (23.75,16.5) circle (0.75cm) node {\Huge $x_{T}$} ;
\draw [ fill={rgb,255:red,0; green,0; blue,0} ] (16.5,16.5) circle (0cm);
\draw  (17.5,16.5) circle (0.25cm);
\draw  (18.5,16.5) circle (0.25cm);
\draw [->, >=Stealth, >={Stealth[length=8pt, width=10pt]}] (6.25,17.25) -- (6.25,18.3);
\draw  (5.6,19.6) rectangle  node {\Huge $s_{\theta}$} (6.9,18.3);
\draw  (10.6,19.6) rectangle  node {\Huge $s_{\theta}$} (11.9,18.3);
\draw  (23.1,19.6) rectangle  node {\Huge $s_{\theta}$} (24.4,18.3);
\draw  (3.75,19) circle (0.75cm) node {\huge $h_{0}$} ;
\draw [->, >=Stealth, >={Stealth[length=8pt, width=10pt]}] (4.5,19) -- (5.58,19);
\draw  (8.75,19) circle (0.75cm) node {\huge $h_{1}$} ;
\node [font=\Huge] at (5.5,18.5) {};
\draw [->, >=Stealth, >={Stealth[length=8pt, width=10pt]}] (6.9,19) -- (8,19);
\node [font=\Huge] at (8.25,18.5) {};
\draw  (13.75,19) circle (0.75cm) node {\huge $h_{2}$} ;
\node [font=\Huge] at (13.25,18.5) {};
\draw  (21.25,19) circle (0.75cm) node {\huge $h_{T-1}$} ;
\draw [->, >=Stealth, >={Stealth[length=8pt, width=10pt]}] (22,19) -- (23.1,19);
\node [font=\Huge] at (20.75,18.5) {};
\draw [ color={rgb,255:red,245; green,116; blue,112}, ->, >=Stealth, >={Stealth[length=8pt, width=10pt]}] (8.6,21.25) -- (8.6,22.25);
\draw [ color={rgb,255:red,245; green,116; blue,112}, ->, >=Stealth, >={Stealth[length=8pt, width=10pt]}] (5.5,20.75) -- (5.5,21.75);
\draw [ color={rgb,255:red,245; green,116; blue,112}, ->, >=Stealth, >={Stealth[length=8pt, width=10pt]}] (7.05,21) -- (7.05,22);
\draw [ color={rgb,255:red,245; green,116; blue,112} ] (4.9,21.75) rectangle  node {\Huge $f_{\theta}^{1}$} (6.1,23.);
\draw [ color={rgb,255:red,245; green,116; blue,112} ] (6.45,22) rectangle  node {\Huge $f_{\theta}^{k}$} (7.65,23.25);
\draw [ color={rgb,255:red,245; green,116; blue,112} ] (8,22.27) rectangle  node {\Huge $f_{\theta}^{K}$} (9.2,23.48);
\draw [ color={rgb,255:red,245; green,116; blue,112}, ->, >=Stealth, >={Stealth[length=8pt, width=10pt]}] (5.5,23) -- (5.5,23.75);
\draw [ color={rgb,255:red,245; green,116; blue,112}, ->, >=Stealth, >={Stealth[length=8pt, width=10pt]}] (7.05,23.25) -- (7.05,24);
\draw [ color={rgb,255:red,245; green,116; blue,112}, ->, >=Stealth, >={Stealth[length=8pt, width=10pt]}] (8.6,23.5) -- (8.6,24.25);
\draw [ color={rgb,255:red,245; green,116; blue,112} ] (5.5,24.5) circle (0.75cm) node {\Huge $\hat{x}_2^{1}$} ;
\draw [ color={rgb,255:red,245; green,116; blue,112} ] (7.05,24.75) circle (0.75cm) node {\Huge $\hat{x}_{2}^k$} ;
\draw [ color={rgb,255:red,245; green,116; blue,112} ] (8.6,25) circle (0.75cm) node {\Huge $\hat{x}_{2}^{K}$} ;
\draw [short] (8.75,19.75) -- (8.75,20.8);
\draw [ color={rgb,255:red,96; green,165; blue,199}, ->, >=Stealth, >={Stealth[length=8pt, width=10pt]}] (13,20.75) -- (13,21.75);
\draw [ color={rgb,255:red,96; green,165; blue,199}, ->, >=Stealth, >={Stealth[length=8pt, width=10pt]}] (9.9,20.25) -- (9.9,21.25);
\draw [ color={rgb,255:red,96; green,165; blue,199}, ->, >=Stealth, >={Stealth[length=8pt, width=10pt]}] (11.45,20.5) -- (11.45,21.5);
\draw [ color={rgb,255:red,96; green,165; blue,199} ] (9.3,21.25) rectangle  node {\Huge $\gamma_{\theta}^{1}$} (10.5,22.5);
\draw [ color={rgb,255:red,96; green,165; blue,199} ] (10.85,21.5) rectangle  node {\Huge $\gamma_{\theta}^{k}$} (12.05,22.75);
\draw [ color={rgb,255:red,96; green,165; blue,199} ] (12.4,21.75) rectangle  node {\Huge $\gamma_{\theta}^{K}$} (13.6,23.0);
\draw [ color={rgb,255:red,96; green,165; blue,199}, ->, >=Stealth, >={Stealth[length=8pt, width=10pt]}] (9.9,22.5) -- (9.9,23.10);
\draw [ color={rgb,255:red,96; green,165; blue,199}, ->, >=Stealth, >={Stealth[length=8pt, width=10pt]}] (11.45,22.75) -- (11.45,23.35);
\draw [ color={rgb,255:red,96; green,165; blue,199}, ->, >=Stealth, >={Stealth[length=8pt, width=10pt]}] (13,23) -- (13,23.6);
\draw [ color={rgb,255:red,96; green,165; blue,199} ] (9.9,23.85) circle (0.75cm) node {\Huge $\hat{\gamma}_{1}$} ;
\draw [ color={rgb,255:red,96; green,165; blue,199} ] (11.45,24.1) circle (0.75cm) node {\Huge $\hat{\gamma}_{k}$} ;
\draw [ color={rgb,255:red,96; green,165; blue,199} ] (13,24.35) circle (0.75cm) node {\Huge $\hat{\gamma}_{K}$} ;
\draw [short] (11.45,20.5) -- (7.05,21);
\draw [ color={rgb,255:red,96; green,165; blue,199}, short] (13,20.75) -- (9.9,20.25);
\draw [ color={rgb,255:red,245; green,116; blue,112}, short] (8.6,21.25) -- (5.5,20.75);
\draw  (26.25,19) circle (0.75cm) node {\huge $h_{T}$} ;
\node [font=\huge] at (25.75,18.5) {};
\draw [->, >=Stealth, >={Stealth[length=8pt, width=10pt]}] (24.4,19) -- (25.5,19);
\draw [->, >=Stealth, >={Stealth[length=8pt, width=10pt]}] (11.9,19) -- (13,19);
\draw [->, >=Stealth, >={Stealth[length=8pt, width=10pt]}] (14.5,19) -- (15.5,19);
\draw [->, >=Stealth, >={Stealth[length=8pt, width=10pt]}] (9.5,19) -- (10.6,19);
\draw [->, >=Stealth, >={Stealth[length=8pt, width=10pt]}] (11.25,17.25) -- (11.25,18.3);
\node [font=\huge] at (7.25,16.75) {};
\draw [->, >=Stealth, >={Stealth[length=8pt, width=10pt]}] (23.75,17.25) -- (23.75,18.3);
\draw [->, >=Stealth, >={Stealth[length=8pt, width=10pt]}] (8.75,19.75) -- (8.75,20.8);
\draw [ color={rgb,255:red,245; green,116; blue,112}, ->, >=Stealth, >={Stealth[length=8pt, width=10pt]}] (21.1,21.25) -- (21.1,22.25);
\draw [ color={rgb,255:red,245; green,116; blue,112}, ->, >=Stealth, >={Stealth[length=8pt, width=10pt]}] (18,20.75) -- (18,21.75);
\draw [ color={rgb,255:red,245; green,116; blue,112}, ->, >=Stealth, >={Stealth[length=8pt, width=10pt]}] (19.55,21) -- (19.55,22);
\draw [ color={rgb,255:red,245; green,116; blue,112} ] (17.4,21.75) rectangle  node {\Huge $f_{\theta}^{1}$} (18.6,23.);
\draw [ color={rgb,255:red,245; green,116; blue,112} ] (18.95,22) rectangle  node {\Huge $f_{\theta}^{k}$} (20.15,23.25);
\draw [ color={rgb,255:red,245; green,116; blue,112} ] (20.5,22.27) rectangle  node {\Huge $f_{\theta}^{K}$} (21.7,23.48);
\draw [ color={rgb,255:red,245; green,116; blue,112}, ->, >=Stealth, >={Stealth[length=8pt, width=10pt]}] (18,23) -- (18,23.75);
\draw [ color={rgb,255:red,245; green,116; blue,112}, ->, >=Stealth, >={Stealth[length=8pt, width=10pt]}] (19.55,23.25) -- (19.55,24);
\draw [ color={rgb,255:red,245; green,116; blue,112}, ->, >=Stealth, >={Stealth[length=8pt, width=10pt]}] (21.10,23.5) -- (21.10,24.25);
\draw [ color={rgb,255:red,245; green,116; blue,112} ] (18,24.5) circle (0.75cm) node {\Huge $\hat{x}_T^{1}$} ;
\draw [ color={rgb,255:red,245; green,116; blue,112} ] (19.55,24.75) circle (0.75cm) node {\Huge $\hat{x}_{T}^k$} ;
\draw [ color={rgb,255:red,245; green,116; blue,112} ] (21.1,25) circle (0.75cm) node {\Huge $\hat{x}_{T}^{K}$} ;
\draw [->, >=Stealth, >={Stealth[length=8pt, width=10pt]}] (21.25,19.75) -- (21.25,20.8);
\draw [ color={rgb,255:red,96; green,165; blue,199}, ->, >=Stealth, >={Stealth[length=8pt, width=10pt]}] (25.5,20.75) -- (25.5,21.75);
\draw [ color={rgb,255:red,96; green,165; blue,199}, ->, >=Stealth, >={Stealth[length=8pt, width=10pt]}] (22.4,20.25) -- (22.4,21.25);
\draw [ color={rgb,255:red,96; green,165; blue,199}, ->, >=Stealth, >={Stealth[length=8pt, width=10pt]}] (23.95,20.5) -- (23.95,21.5);
\draw [ color={rgb,255:red,96; green,165; blue,199} ] (21.8,21.25) rectangle  node {\Huge $\gamma_{\theta}^{1}$} (23,22.5);
\draw [ color={rgb,255:red,96; green,165; blue,199} ] (23.35,21.5) rectangle  node {\Huge $\gamma_{\theta}^{k}$} (24.55,22.75);
\draw [ color={rgb,255:red,96; green,165; blue,199} ] (24.90,21.75) rectangle  node {\Huge $\gamma_{\theta}^{K}$} (26.1,23.0);
\draw [ color={rgb,255:red,96; green,165; blue,199}, ->, >=Stealth, >={Stealth[length=8pt, width=10pt]}] (22.4,22.5) -- (22.4,23.10);
\draw [ color={rgb,255:red,96; green,165; blue,199}, ->, >=Stealth, >={Stealth[length=8pt, width=10pt]}] (23.95,22.75) -- (23.95,23.35);
\draw [ color={rgb,255:red,96; green,165; blue,199}, ->, >=Stealth, >={Stealth[length=8pt, width=10pt]}] (25.5,23) -- (25.5,23.6);
\draw [ color={rgb,255:red,96; green,165; blue,199} ] (22.4,23.85) circle (0.75cm) node {\Huge $\hat{\gamma}_{1}$} ;
\draw [ color={rgb,255:red,96; green,165; blue,199} ] (23.95,24.1) circle (0.75cm) node {\Huge $\hat{\gamma}_{k}$} ;
\draw [ color={rgb,255:red,96; green,165; blue,199} ] (25.5,24.35) circle (0.75cm) node {\Huge $\hat{\gamma}_{K}$} ;
\draw [short] (23.95,20.5) -- (19.55,21);
\draw [ color={rgb,255:red,96; green,165; blue,199}, short] (25.5,20.75) -- (22.4,20.25);
\draw [ color={rgb,255:red,245; green,116; blue,112}, short] (21.1,21.25) -- (18,20.75);

\draw [ fill={rgb,255:red,0; green,0; blue,0} ] (16.5,16.5) circle (0cm);
\draw [ line width=0.3pt ] (16.5,16.5) circle (0.25cm);
\draw [ fill={rgb,255:red,0; green,0; blue,0} ] (16.5,19) circle (0cm);
\draw  (17.5,19) circle (0.25cm);
\draw  (18.5,19) circle (0.25cm);
\draw [ fill={rgb,255:red,0; green,0; blue,0} ] (16.5,19) circle (0cm);
\draw [ line width=0.3pt ] (16.5,19) circle (0.25cm);
\draw [->, >=Stealth, >={Stealth[length=8pt, width=10pt]}] (19.5,19) -- (20.5,19);
\end{circuitikz}
}
    \caption{\textbf{Components of \texttt{TimeMCL.}} Prediction heads are in \textcolor{mylightcoral}{Lightcoral} color, and Score heads are in \textcolor{myblue}{Blue}. Rectangles contain functions, and circles contain features.}
    \label{fig:tikz_example}
\end{figure}

This two-step optimization allows to optimize the loss in an alternating fashion, despite the non-differentiability of the $\mathrm{min}$ operator. Note that the latter can be computed batch-wise on the historical data, computing the Winner head of each batch index, with a loss that writes as:
\begin{equation}
\mathcal{L}^{\mathrm{WTA}}(\theta_1,\dots, \theta_{K}) \triangleq \mathbb{E}_{\xTotal}[\min_{k=1,\dots,K} \Lk(\xPast,\xFuture)]\;,
\label{eq:wta_loss_bw}
\end{equation}
where the expectation is taken over $(\xPast, \xFuture) \sim p(\xTotal)$. 
While the WTA Loss trains several models with the aim of producing several trajectories, we also use \textit{score} heads as in  \citet{letzelter2023resilient} $\gamma_{\theta}^{1},\dots,\gamma_{\theta}^{K}: \mathcal{H} \rightarrow [0,1]$ to learn to predict the probability of head being the winner and avoid overconfident heads. The latter are trained with
\begin{equation}
\mathcal{L}^{\mathrm{s}} \triangleq 
\mathbb{E}_{\xTotal} \left[
\sum_{\substack{k=1,\dots,K \\ t = t_{0},\dots,T}} 
\operatorname{BCE} \left(\mathds{1}\left[k = k^{\star}\right], \gamma_\theta^k(h_{t-1}) \right)
\right]\,,
\end{equation}
where the binary cross entropy $\mathrm{BCE}(p, q) \triangleq-p \log (q)-$ $(1-p) \log (1-q)$, aligns the predicted and target winner assignation probabilities. The final training objective is a compound loss $\mathcal{L} = \mathcal{L}^{\mathrm{WTA}} + \beta \mathcal{L}^{\mathrm{s}}$, where $\beta > 0$ is the confidence loss weight. See Figure \ref{fig:tikz_example} for an illustration of the components of \texttt{TimeMCL}.

\subsection{Inference and Sampling}
\label{subsec:inference}

Once trained, \texttt{TimeMCL} provides $K$ plausible predictions $\hat{x}^{1} \sim p_{\theta}^{1}, \dots, \hat{x}^{K} \sim p_{\theta}^{K}$, by ancestral sampling on each of the models. When using recurrent neural networks, the unrolling procedure described in Section \ref{sec:setup} can be applied by first encoding the input sequence $s_{\theta}$ to obtain a hidden state $h_{t_{0}}$, and unrolling the autoregressive model. As in Section \ref{sec:setup}, we encapsulate these operation with unrolled networks $\mathscr{F}_{\theta}^1, \dots, \mathscr{F}_{\theta}^K$.
The scores are computed in the same way, using the score heads $\gamma_{\theta}^k$ instead of the prediction heads $f_{\theta}^k$, and we denote $\Gamma_{\theta}^k$ their unrolled networks. To get a single score associated with each predicted trajectory, we averaged the predicted scores over the sequence.

In cases where the ambiguity is reduced, such that for short-horizon forecasts when only one prediction is required, one might pick the best head according to the predicted score or sample from them in proportion to some confidence measure. In a longer autoregressive forecast, we can consider the $K$ outputs at each time step ---thus producing a set of possible futures from the single forward pass.

\subsection{Taking advantage of Winner-takes-all variants}

While the WTA Loss has been proven effective for handling ambiguous tasks \cite{lee2016stochastic, seo2020trajectory, garcia2021distillation}, some heads may theoretically be under-trained \cite{rupprecht2017learning}. This may occur when a single mode (or a few modes) dominates the target distribution, or due to suboptimal initialization, similar to what can happen in Lloyd’s algorithm for $K$-Means clustering \cite{lloyd1982least, arthur2007k}. In this case, the scoring loss $\mathcal{L}^{\mathrm{s}}$ ensures setting a low probability to those concerning hypotheses so that the latter can be ignored at inference.

It is possible to mitigate this issue, and therefore improve the performance of the estimator by using \emph{relaxation} techniques of the $\min$ operator. In this case, the best head loss in \eqref{eq:wta_loss_bw} can be substituted with weighted loss from the different heads:
\begin{equation}
    \tilde{\mathcal{L}}^{\mathrm{WTA}}(\theta_1,\dots,\theta_k) \triangleq \mathbb{E}_{\xTotal}\left[\sum_{k=1}^{K} q_{k} \Lk (\xPast, \xFuture) \right]\;,
\end{equation}

where the coefficients $q_{k} \geq 0$ sum to one, and allow to assign some weight to non-winner hypotheses.

This idea was originally suggested through Relaxed Winner-takes-all (R-WTA) loss proposed by \citet{rupprecht2017learning}, which suggested back-propagating not only on the winning head $k^{\star}$, but also on the non-winner. In this case $q_{k^{\star}} = 1 - \varepsilon$ and $q_k = \frac{\varepsilon}{K-1}$ for $k \neq k^{\star}$ (See \eqref{eq:rwta_loss} in Appendix \ref{apx:baselines}). More recently, \citet{perera2024annealed} proposed an annealed method inspired from Deterministic annealing \cite{rose1990deterministic} using a \textit{softmin} operator:
\begin{equation}\label{eq:awta_assignation}
q_{k}(T) = \frac{1}{Z(\xPast,\xFuture;T)} \exp \Bigl(-\frac{\Lk(\xPast,\xFuture)}{T}\Bigr),\end{equation}
with $$Z(\xPast,\xFuture;T) \triangleq \sum_{s=1}^{K} \exp \Bigl(-\frac{\Ls(\xPast,\xFuture)}{T}\Bigr)\;,$$
where the temperature $T$ is annealed during training, \textit{e.g.,} considering at training step $t$, $T(t) = T(0) \rho^{t}$ with $\rho < 1$. At higher temperatures, the target assignment is effectively softened, making the early stages of training easier.

We implemented these two variants, within our \texttt{TimeMCL} model in the \texttt{Gluonts} framework \cite{alexandrov2020gluonts}. Based on our experience, these methods are meaningful variations of WTA that were worth exploring, as they demonstrated improvements over vanilla WTA in certain configurations. 

\section{Theoretical analysis and interpretation}

In this section, based on the notion of functional quantization, we provide insights into the interpretation of our approach. In particular, we show that, under squared error, the $K$ heads form a \emph{Voronoi} tessellation of future trajectories 
and act as a \emph{conditional codebook}. This viewpoint explains how WTA theoretically captures the best possible way the conditional law over stochastic processes, given a sampling budget of $K$ predictions. Our claims are then illustrated through a synthetic data example, specifically focusing on certain Gaussian Processes.

\subsection{\texttt{TimeMCL} is a stationary conditional functional quantizer}

For simplicity and without loss of generality, let us temporarily assume that $\beta = 0$, \textit{i.e.,} only the WTA Loss $\mathcal{L}^{\mathrm{WTA}}$ is optimized. When predicting the future of a time series given its context $\xPast$, we are effectively observing, during supervision, a (partial) path realization $\xFuture$ of an underlying stochastic process. Following standard functional data analysis \cite{bosq2000linear,ramsay2005functional}, we assume that each time-series trajectory $x_{t_0:T}$ belongs to a Banach space $(E, \|\cdot\|)$.
For concreteness, one may consider $E = L^2([t_0,T])$ endowed with the usual $L^2$ norm,
though any separable Banach space is admissible for the theoretical arguments to hold. In this space, the distance $d(\cdot,\cdot)$ induced by $\|\cdot\|$ 
enables us to define the Voronoi tessellation over future paths $x_{t_0:T}$ 
as described below. 

The $L^{2}$ quantization problem aims at finding the best approximation of a random vector $X$, using $K$ points in $E$. The quality of the approximation using at most $K$ points $\{f_1,\dots,f_K\}$, is generally measured with the distortion, defined as:
$D_{2}(X) = \mathbb{E}_{X} [ \mathrm{min}_{f \in \{f_1,\dots,f_K\}}\;d(X,f)^2]\;,$ which is finite over if $X$ admits a second order moment. Note that our learning setup involves quantization in \textit{conditional} form, \textit{i.e.,} the random variable of interest depends on a context.

We state hereafter our main theoretical result, showing that \texttt{TimeMCL} can provably perform functional quantization of the target space of plausible trajectories. It can be seen as an adaptation of Proposition 5.2 in \citet{letzelter24winner} to the case of functional quantization of time series.

\begin{proposition} 
\label{prop:opt_quantizer}
See Proposition \ref{prop:centroidal_prop_apx} in the Appendix. Under the assumptions that:
\begin{enumerate}
    \item The batch size is big enough so that the difference between the $\mathcal{L}^{\mathrm{WTA}}$ risk and its empirical version can be neglected (Assumption \ref{asm:true_risk}). 
    \item The neural network we are considering is expressive enough so that minimizing the risk is equivalent to minimizing the input-dependent risk for each context $\xPast$ (Assumption \ref{asm:expressiveness}). 
    \item The training has converged and $\mathcal{L}^{\mathrm{WTA}}$ has reached a local minima (Assumption \ref{asm:optimality}).
\end{enumerate}
Then, \texttt{TimeMCL} is a conditional stationary quantizer for each sampled window $(\xPast, \xFuture)$, that is, for each $k \in [\![1,K]\!]$:
\begin{equation}
\mathscr{F}_{\theta}^{k}(\xPast) = \mathbb{E}[\xFuture \mid x_{t_0:T} \in \Yk(\xPast)]\;,
\label{eq:centroidal}
\end{equation}
where \[
\Yk(\xPast) 
\,=\,
\Bigl\{
\xFuture \mid 
\Lk <
\Lr
\;, \forall r\neq k
\Bigr\}\;.
\]
We denoted by abuse of notations $\Lk = \Lk(\xPast, \xFuture)$ for simplicity and the same for $\Lr$. This makes \texttt{TimeMCL} akin to a conditional and gradient-based version of K-Means over the set of possible future trajectories. 
\end{proposition}
\begin{proof}[Sketch of proof.] See Proposition \ref{prop:centroidal_prop_apx} for the full proof. The demonstration of this result is made by first leveraging Assumption \ref{asm:true_risk} to re-write the WTA Loss in the form
\begin{equation}
\label{eq:risk}
\mathbb{E}_{\xPast} \left[ \sum_{k=1}^{K} \int_{\Yk(\xPast)} \Lk \; \mathrm{d}p(\xFuture \mid \xPast) \right]\;,
\end{equation}
where each $\xFuture \in \Yk(\xPast)$ picks the head $k$ it is closest to. Under the expressivity assumption (Assumption \ref{asm:expressiveness}), \eqref{eq:risk} comes down to optimizing a functional \eqref{eqapx:functional} of the hypotheses position for each fixed context $\xPast$. We assume that, during training, our predictor generates trajectories solely from the context (\textit{i.e.,} independently of the observed values), effectively as though $\mathscr{F}_{\theta}^1, \dots, \mathscr{F}_{\theta}^K$ were used directly for training.

From here, we leverage the fact that \texttt{TimeMCL} is a two-step training procedure, and then the alternating optimization argument from \citet{rupprecht2017learning} (Theorem 1) applies to obtain the optimal centroids. This is performed using $L^2$ square loss for $\ell$, from the vanishing gradient condition on the optimized functional (Assumption \ref{asm:optimality}). We also say that in this case, Voronoi tesselation on the trajectory space induced by the hypothesis is centroidal \cite{du1999centroidal}. 
\end{proof}

Now if we assume $\beta > 0$, we can show the following proposition.

\begin{proposition}\label{prop:opt_quantizer_scores}
Under similar assumptions as in Proposition \ref{prop:opt_quantizer}, one can show that a necessary optimality condition for the score heads is that \begin{equation}
\label{eq:scores}
\Gamma_\theta^k\left(\xPast\right) = \mathbb{P} \left(\xFuture \in \Yk\left( \xPast\right) \mid \xPast \right).
\end{equation}

\end{proposition}
\begin{proof}
    Full proof in Appendix \ref{proof:unbiaised_estimator_voronoi_cell}. 
\end{proof} 

\texttt{TimeMCL} can thus be viewed as a \emph{conditional vector quantization} scheme \cite{gersho1992vector}, 
where each head $k$ is a \emph{code vector} (in functional form). 

By conditioning on the past data $\xPast$, \texttt{TimeMCL} effectively learns a family of partitions in the time-series trajectory space. If the number of hypotheses is large, increasing $K$ under the above assumptions typically reduces reconstruction error at a rate akin to $K^{-2/d}$, in line with classical quantization theory \cite{zador1982asymptotic}.

\subsection{Smoothness of \texttt{TimeMCL} predictions for (most) time series}
\label{sec:smoothness}

As we have shown, under certain assumptions, if the model reaches a stationary point,  
\[
\mathscr{F}_\theta^k\left(x_{1: t_0-1}\right) = \mathbb{E}\left[x_{t_0: T} \mid x_{t_0: T} \in \mathcal{X}_k(\xPast)\right]\;,
\]  
and we can interpret the prediction as a mean of different hypothetical trajectories. Since most stochastic time series contain centered noise—caused by various factors such as measurement errors or random events—the averaging process tends to eliminate this noise, resulting in a smooth appearance. We observe this phenomenon consistently in the real examples we visualized. We also noticed that the appearance of smoothness increases as training progresses. This property of \textit{mean} predictions reinforces our conviction that the model is providing representative trajectories rather than just random samples.

\subsection{Synthetic data example}
\label{sec:synthetic}
\begin{figure*}[t]
    \centering
    \includegraphics[width=\linewidth]{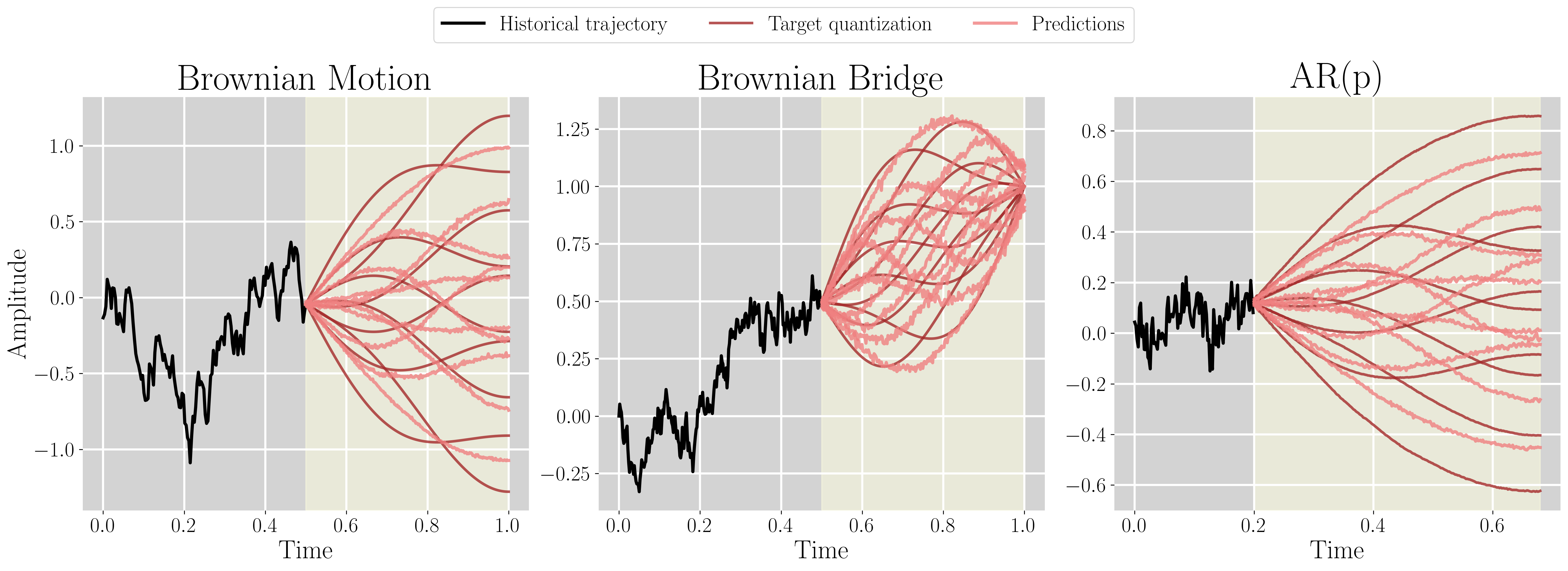}
    \caption{\textbf{Conditional Quantization of Stochastic Processes with \texttt{TimeMCL}.} The Figure shows the predictions of \texttt{TimeMCL} on three synthetic datasets as described in Section \ref{sec:synth} and Appendix \ref{appendix:expr_synth}. Predictions of \texttt{TimeMCL} are shown in \textcolor{mylightcoral}{lightcoral} color, and target quantization in each case are shown in \textcolor{mybrown}{brown}. We used $10$ hypotheses here, a three-layer MLP as backbone with score-heads disabled in those toy experiments. Brownian Motion, Brownian Bridge, and AR(p) are increasingly complex in terms of conditioning dependencies in the context window (See Section \ref{sec:synth}). We see that in those three cases, \texttt{TimeMCL} predictions nicely approximate the shape of target quantization functions, justifying its interpretation as a \textit{conditional functional quantizer}.}
    \label{fig:toy}
\end{figure*}

\label{sec:synth}
We evaluate \texttt{TimeMCL} on three synthetic processes: Brownian motion, a Brownian bridge, and an AR(5) process. While the Brownian motion serves as a simple example with minimal context dependence, the Brownian bridge introduces a time-conditioned structure, and the AR(5) process tests the model’s ability to handle stronger context dependencies. Training is performed on randomly sampled trajectories with appropriate conditioning, and quantization is assessed using theoretical references: Karhunen-Loève-based quantization for Brownian motion and the Brownian bridge, and Lloyd’s algorithm for the AR(5) process (Appendix~\ref{appendix:expr_synth}).  

For this toy experiment, we used a three-layer MLP that predicts the entire sequence at once from its context (see Appendix \ref{appendix:expr_synth} for details). To keep the implementation as straightforward as possible, we kept the model's parameterization lightweight and omitted the score heads.

As shown in Figure \ref{fig:toy}, \texttt{TimeMCL} consistently produces smooth and near-optimal quantizations across all settings. For Brownian motion, the predicted trajectories closely align with the theoretical optimal quantization, demonstrating the model’s ability to learn conditional distributions from minimal context. In the Brownian bridge setting, where the conditioning depends on both time and value, the model successfully captures the structural constraints, leading to coherent and well-quantized trajectories. The AR(5) process presents a more complex challenge due to its stronger temporal dependencies, yet \texttt{TimeMCL} effectively utilizes the past observations to produce long-horizon predictions that remain consistent with Lloyd’s quantization. The results highlight the model’s ability to condition on past observations, effectively adapting to different processes and maintaining predictive stability over extended time horizons.  

\section{Empirical evaluation}

In this section, we empirically validate our method, with experiments on real-world time series. Compared to the experiments in Section \ref{sec:synth}, the underlying law of the data-generating process is not known, making the task more realistic. The goal is to compare \texttt{TimeMCL} with state-of-the-art probabilistic time series forecasters, emphasizing its balance between quantization, predictive performance, and computational efficiency.

\begin{figure*}[t]
    \centering
    \includegraphics[width=1.\linewidth]{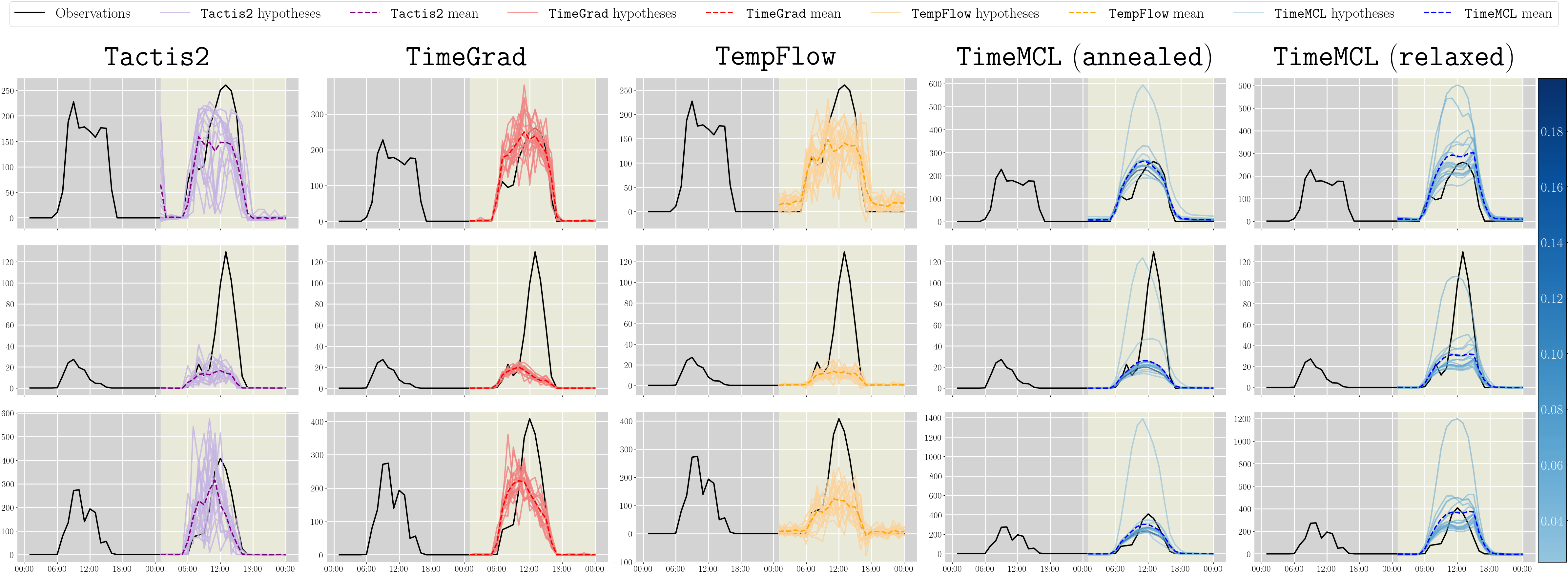}
    \caption{\textbf{Qualitative results.} Visualization of the predictions on the \textsc{Solar} dataset, comparing \texttt{Tactis2} \textcolor{mypurple}{(purple)}, \texttt{TimeGrad} \textcolor{red}{(red)}, \texttt{TempFlow} \textcolor{orange}{(orange)}, and \texttt{TimeMCL} trained with two relaxation techniques \textcolor{myblue}{(blue)}. Each model predicts sixteen hypotheses, dashed lines represent the predicted mean of the conditional distribution which was computed as the weighted sum of the predicted hypotheses by the scores for \texttt{TimeMCL}. Each row represents a different dimension ($\dimn = 137$ here). The light yellow zone highlights the prediction window, and the score intensities are displayed as shaded blue lines, as per the scale on the right. \texttt{DeepAR} is not included in this visualization as it does not perform competitively. In this example, \texttt{TimeGrad}, \texttt{TempFlow} and \texttt{Tactis2} generate meaningful hypotheses. However, \texttt{TimeMCL}’s predictions are noticeably smoother and more diverse.}
    \label{fig:combined_plot_main_paper}
\end{figure*}

\subsection{Experimental setup}

\textbf{Datasets.} Our approach is evaluated on six well-established benchmark datasets taken from \texttt{Gluonts} library \cite{alexandrov2020gluonts}, preprocessed exactly as in \citet{salinas2019high, rasul2021autoregressive}. Each dataset consists of strictly positive and bounded real-valued time series. The characteristics of these datasets are summarized in Table \ref{tab:datasets_characteristics} in the Appendix. \textsc{Solar} contains hourly solar power outputs from 137 sites with strong daily seasonality. \textsc{Electricity} records hourly power consumption for 370 clients, exhibiting daily and weekly periodicities. \textsc{Exchange} tracks daily exchange rates for eight currencies, showing less seasonality and being influenced by macroeconomic factors. \textsc{Traffic} provides hourly occupancy rates from 963 road sensors, capturing rush-hour peaks and weekly patterns. \textsc{Taxi} comprises time series of NYC taxi rides recorded at 1,214 locations. \textsc{Wikipedia} includes daily views of 2,000 Wikipedia pages \cite{gasthaus2019probabilistic}.  
See \citet{lai2018modeling} and Appendix \ref{apx:datasets} for an extensive description.

\textbf{Metrics.} We evaluate our approach with six different metrics, each of them being detailed in Appendix \ref{apx:metrics}. First, we considered the Distortion, which is computed as $$D_{2} = \frac{1}{N} \sum_{i=1}^{N} \min_{k=1,\dots,K} d\left(\mathscr{F}^k_{\theta}(\xPast^i),\xFuture^i\right)
\;,$$
consisting of computing the Euclidean distance $d$ between the target series and closest hypothesis, averaged over the test set with $N$ samples. This allows us to compare fairly with other baselines when the sample size $K$ is fixed. To assess computational efficiency during inference, we report both the inference Floating Point Operations (FLOPs) and run-time for each baseline in Table \ref{tab:flops}. As a means to validate the theoretical claim presented in Section \ref{sec:smoothness}, we also compute the Total Variation (TV), which quantifies the smoothness of the predicted trajectories in Table \ref{tab:TV Score}. Finally, for comprehensive comparison, we include the Average Root Mean Square Error (RMSE) and the Continuous Ranked Probability Score (CRPS) (summed over all time series dimensions). These results are reported in Tables \ref{tab:rmse_sum} and \ref{tab:crps_sum} in the Appendix.

\textbf{Baselines.} We considered the following baselines: \texttt{ETS} \cite{hyndman2008forecasting}, \texttt{Tactis2} \cite{ashok2024tactis}, \texttt{DeepAR} \cite{salinas2020deepar}, \texttt{TempFlow} \cite{rasul2021multivariate}, and \texttt{TimeGrad} \cite{rasul2021autoregressive}. These were compared against \texttt{TimeMCL} with two relaxation techniques: Relaxed-WTA \cite{rupprecht2017learning} and aMCL \cite{perera2024annealed}. Note that both of these Multiple Choice Learning variants use score heads as in \citet{letzelter2023resilient} (with $\beta=0.5$).

\textbf{Architectures.} We compare \texttt{DeepAR}, \texttt{TempFlow}, and \texttt{TimeGrad} with \texttt{TimeMCL}, using the same neural network backbone: an RNN with LSTM cells, as in the original implementations \cite{hochreiter1997long}. This ensures fair comparison conditions. In these experiments, each hypothesis head in \texttt{TimeMCL} and the projection layer of \texttt{DeepAR} consists of a single linear layer. Meanwhile, the noise prediction in \texttt{TimeGrad} is implemented with a dilated ConvNet featuring residual connections \cite{van2016wavenet, kong2021diffwave, rasul2021autoregressive}. Additionally, we include comparisons with methods based on transformer backbones, such as the transformer-based version of \texttt{TempFlow}, (named \texttt{Trf.TempFlow} \cite{rasul2021multivariate}) and \texttt{Tactis2} \cite{drouin2022tactis, ashok2024tactis}, which leverages copulas for modeling dependencies. Note that \texttt{ETS} does not use a neural network.

\textbf{Training.} Training is conducted using the Adam optimizer with an initial learning rate of $10^{-3}$ for $200$ training epochs. During each epoch, $30$ batches of size $200$ are sampled from the historical data, considering random windows with a context set equal to the prediction length. We used a validation split of size equal to 10 times the prediction length. Except for \texttt{Tactis2}, which uses Z-Score normalization, the data are scaled by computing the mean value dimension by dimension over the context and dividing the target by this mean. This scaling follows the \texttt{TimeGrad} experimental setup \cite{rasul2021autoregressive}, ensuring consistency. The model is trained on the scaled data, and the inverse transformation is applied later for prediction.

\textbf{Evaluation.} 
The evaluation dataset is divided into multiple non-overlapping subsets, each containing sufficient points for both context and prediction lengths, allowing comprehensive assessment across different temporal segments. To compute \texttt{TimeMCL} metrics while accounting for predicted hypothesis probabilities, we resample with replacement from the $K$ hypotheses obtained in a single forward pass, weighting them by their predicted scores before computing the metrics.

\subsection{Results}

\begin{table*}
    \begin{center}
    \caption{\textbf{Distortion Risk with 16 Hypotheses.} \textbf{\texttt{TimeMCL(R.)}} and \textbf{\texttt{TimeMCL(A.)}} correspond to the relaxed and annealed variants. \textbf{\texttt{ETS}}, \textbf{\texttt{Trf.TempFlow}} and \textbf{\texttt{Tactis2}}, columns are in \textcolor{gray}{gray} because they don't share the same backbone as the other baselines.
    The test scores are averaged over four training seeds for each model. Best scores are in \textbf{bold}, and second-best are \underline{underlined}.}
    \vspace{2pt}
    \label{tab:dist_16hyp}
    \resizebox{2\columnwidth}{!}{
    \begin{tabular}{lcccccc||cc}
\toprule
{}  & \textcolor{gray}{\textbf{\texttt{ETS}}} & \textcolor{gray}{\textbf{\texttt{Trf.TempFlow}}} & \textcolor{gray}{\textbf{\texttt{Tactis2}}} &      \textbf{\texttt{TimeGrad}}  &  \textbf{\texttt{DeepAR}}  &  \textbf{\texttt{TempFlow}}  &  \textbf{\texttt{TimeMCL(R.)}}  &  \textbf{\texttt{TimeMCL(A.)}} \\
\midrule
\textsc{Elec.}    & \textcolor{gray}{14041 $\pm$ 877} & \textcolor{gray}{13519 $\pm$ 2789} & \textcolor{gray}{11616 $\pm$ 2160} &         \textbf{9872 $\pm$ 668}  &         133107 $\pm$ 2870  &             14836 $\pm$ 596  &    \underline{11604 $\pm$ 1042}  &                11611 $\pm$ 1206 \\
\textsc{Exch.}    & \textcolor{gray}{0.051 $\pm$ 0.005} & \textcolor{gray}{0.062 $\pm$ 0.015} & \textcolor{gray}{\textbf{0.030 $\pm$ 0.002}} &               0.038 $\pm$ 0.005  &         0.061 $\pm$ 0.001  &           0.051 $\pm$ 0.007  &   \underline{0.035 $\pm$ 0.005}  &               0.044 $\pm$ 0.005 \\
\textsc{Solar}    & \textcolor{gray}{641.32 $\pm$ 3.57} & \textcolor{gray}{374.69 $\pm$ 20.97} & \textcolor{gray}{358.01 $\pm$ 38.47} &               360.6 $\pm$ 34.79  &        748.68 $\pm$ 18.92  &           371.14 $\pm$ 18.2  &     \textbf{280.03 $\pm$ 15.22}  &    \underline{305.5 $\pm$ 4.05} \\
\textsc{Traffic}  & \textcolor{gray}{2.64 $\pm$ 0.01} & \textcolor{gray}{1.26 $\pm$ 0.02} & \textcolor{gray}{0.84 $\pm$ 0.04} &                 0.78 $\pm$ 0.05  &           2.12 $\pm$ 0.08  &             1.21 $\pm$ 0.04  &        \textbf{0.68 $\pm$ 0.01}  &     \underline{0.72 $\pm$ 0.05} \\
\textsc{Taxi}     & \textcolor{gray}{583.52 $\pm$ 0.41} & \textcolor{gray}{278.22 $\pm$ 13.8} & \textcolor{gray}{243.63 $\pm$ 9.1} &   \underline{209.64 $\pm$ 3.14}  &         407.4 $\pm$ 14.09  &          268.65 $\pm$ 10.69  &      \textbf{187.81 $\pm$ 6.17}  &              229.26 $\pm$ 22.06 \\
\textsc{Wiki}     & \textcolor{gray}{715150 $\pm$ 4742} & \textcolor{gray}{515285 $\pm$ 16000} & \textcolor{gray}{\textbf{254675 $\pm$ 3870}} &   \underline{261361 $\pm$ 1028}  &         368087 $\pm$ 8159  &           382554 $\pm$ 4149  &               261950 $\pm$ 9504  &               267624 $\pm$ 8433 \\
\bottomrule
\end{tabular}
}
    \end{center}
    \end{table*}
\begin{table*}
    \begin{center}
    \caption{\textbf{Computational cost at inference of neural-based methods for $K = 16$ hypotheses on \textsc{Exchange}.} FLOPs are computed with a single batch of size $1$. Run-times are averaged over $15$ random seeds.}
    \label{tab:flops}
    \vspace{2pt}
    \resizebox{1.5\columnwidth}{!}{\begin{tabular}{lccccc||c}
\toprule
 & \textbf{\texttt{Trf.TempFlow}} & \textbf{\texttt{Tactis2}} & \textbf{\texttt{TimeGrad}} & \textbf{\texttt{DeepAR}} & \textbf{\texttt{TempFlow}} & \textbf{\texttt{TimeMCL}} \\
\midrule
\textsc{FLOPs. ($\downarrow$)} & $2.04 \times 10^{8}$ & $1.85 \times 10^{8}$ & $3.05 \times 10^{9}$ & $\mathbf{4.65 \times 10^{5}}$ & $9.29 \times 10^{7}$ & $\underline{8.83 \times 10^{6}}$ \\
\textsc{Run Time. ($\downarrow$)} & $2.47 \pm 0.23$ & $8.69 \pm 0.36$ & $241.57 \pm 2.24$ & $\mathbf{0.70 \pm 0.04}$ & $1.39 \pm 0.03$ & $\underline{1.12 \pm 0.04}$ \\
\bottomrule
\end{tabular}
}
    \end{center}
    \end{table*}

Tables \ref{tab:dist_16hyp} and \ref{tab:flops} show Distortion performance, FLOPs, and run-time, comparing \texttt{TimeMCL} with two tested WTA variants with 16 hypotheses — WTA-Relaxed \cite{rupprecht2017learning} and aMCL \cite{perera2024annealed} — against the baselines. Table \ref{tab:TV Score} displays Total Variation comparison, as a means to quantify smoothness.

\textbf{Distortion and Computation Cost.} Table \ref{tab:dist_16hyp} demonstrates that \texttt{TimeMCL}, particularly when trained with its Relaxed variant, achieves competitive performance compared to the other models when the number of hypotheses is fixed (with $K = 16$ here). This is especially promising, as \texttt{Tactis2} and \texttt{TimeGrad}, which are the strongest competitors in terms of distortion, incur significantly higher FLOPs and run-time (see Table \ref{tab:flops}). A similar trend is observed in Table \ref{tab:dist_8hyp} in the Appendix, which shows results for $8$ hypotheses. This behavior is expected since \texttt{TimeMCL} explicitly optimizes for distortion. It’s worth noting that, among the neural-based methods, \texttt{TimeMCL} is the second most computationally efficient model, just behind \texttt{DeepAR}, while achieving significantly better distortion scores. On that account, \texttt{TimeMCL} strikes a promising trade-off between computational cost and performance. For more details, refer to Appendices \ref{secapx:flops} and \ref{secapx:inference_time}.

\textbf{Smoothness.} Table \ref{tab:TV Score} displays the Total Variation defined as $\mathrm{TV} = \frac{1}{K} \sum_{k=1}^{K} \sum_{t=t_{0}}^{T} \left\|\hat{x}_{t+1}^{k}-\hat{x}_{t}^{k}\right\|_2$ which quantifies the average smoothness of the predicted trajectories (lower is more smooth). We see that \texttt{TimeMCL}, when trained either with the annealed or relaxed variant, provides significantly smoother trajectories compared to the baselines, further confirming the claim of Section \ref{sec:smoothness} as a consequence of Proposition \ref{prop:opt_quantizer}.

\textbf{Comparing  \texttt{TimeMCL} with the baselines on standard metrics.} Table \ref{tab:rmse_sum} and \ref{tab:crps_sum} provides performance on CRPS and RMSE, respectively. We see that, except for CRPS and RMSE on \textsc{exchange} (for which \texttt{Tactis} is better), \texttt{TimeGrad} outperforms the other baselines. We also observe that  \texttt{TimeMCL} is competitive despite optimizing a completely different training objective, at a fraction of \texttt{TimeGrad}’s and \texttt{Tactis}’s computational cost.

\begin{table}
    \begin{center}
    \caption{\textbf{Total Variation ($\downarrow$) comparison for $K = 16$ hypotheses.} See Table \ref{tabapx:TV Score} for the full scores.}
    \label{tab:TV Score}
    \vspace{-2pt}
    \resizebox{\columnwidth}{!}{\begin{tabular}{lccc||cc}
\toprule
{} &          \textbf{\texttt{Tactis2}} &         \textbf{\texttt{TimeGrad}} &          \textbf{\texttt{TempFlow}} &                \textbf{\texttt{TimeMCL(R.)}} &                \textbf{\texttt{TimeMCL(A.)}} \\
\midrule
\textsc{Elec.}   &    284232 \scriptsize{$\pm$ 23269} &    372443 \scriptsize{$\pm$ 18499} &     434097 \scriptsize{$\pm$ 34623} &      \textbf{220375 \scriptsize{$\pm$ 33961}} &   \underline{245908 \scriptsize{$\pm$ 30505}} \\
\textsc{Exch.}   &    0.237 \scriptsize{$\pm$ 0.0355} &    0.606 \scriptsize{$\pm$ 0.0208} &     1.104 \scriptsize{$\pm$ 0.1582} &      \textbf{0.031 \scriptsize{$\pm$ 0.0075}} &   \underline{0.042 \scriptsize{$\pm$ 0.0141}} \\
\textsc{Solar}   &        4421 \scriptsize{$\pm$ 828} &       5383 \scriptsize{$\pm$ 2173} &         3653 \scriptsize{$\pm$ 488} &       \underline{3391 \scriptsize{$\pm$ 947}} &          \textbf{2195 \scriptsize{$\pm$ 297}} \\
\textsc{Traffic} &    11.063 \scriptsize{$\pm$ 0.874} &    12.991 \scriptsize{$\pm$ 1.661} &     18.114 \scriptsize{$\pm$ 0.512} &     \underline{5.766 \scriptsize{$\pm$ 0.19}} &       \textbf{5.714 \scriptsize{$\pm$ 0.283}} \\
\textsc{Taxi}    &  5452.77 \scriptsize{$\pm$ 324.07} &   4232.16 \scriptsize{$\pm$ 713.9} &    4147.2 \scriptsize{$\pm$ 362.87} &  \underline{709.86 \scriptsize{$\pm$ 332.07}} &     \textbf{701.61 \scriptsize{$\pm$ 199.35}} \\
\textsc{Wiki}    &  2634547 \scriptsize{$\pm$ 597323} &  2458503 \scriptsize{$\pm$ 151593} &  12410909 \scriptsize{$\pm$ 319517} &        \textbf{9532 \scriptsize{$\pm$ 10690}} &  \underline{271611 \scriptsize{$\pm$ 359320}} \\
\bottomrule
\end{tabular}
}
    \end{center}
    \end{table}

\textbf{Qualitative comparison.} We qualitatively compare the predictions of \texttt{TimeMCL} with those of the baselines on the \textsc{Solar}, \textsc{Traffic}, and \textsc{Electricity} datasets, as shown in Figure~\ref{fig:combined_plot_main_paper} and in Figures~\ref{fig:qualitative_traffic} and~\ref{fig:qualitative_elec} in the Appendix. Figure \ref{fig:combined_plot_main_paper} demonstrates the ability of the hypotheses to predict multiple futures with the aim of capturing different modes of the target data distribution. \texttt{Tactis2}, \texttt{TimeGrad}, \texttt{TempFlow} struggle to produce predictions that deviate significantly from the mean, with most of their predictions being sampled from the same mode of the distribution. Interestingly, \texttt{TimeMCL} demonstrates the ability to generate predictions far from the mean with a non-negligible probability, indicating that the model successfully captures different modes and estimates the probability of each mode, including possibly rare events.

\begin{figure*}[ht!]
    \centering
    \includegraphics[width=\linewidth]{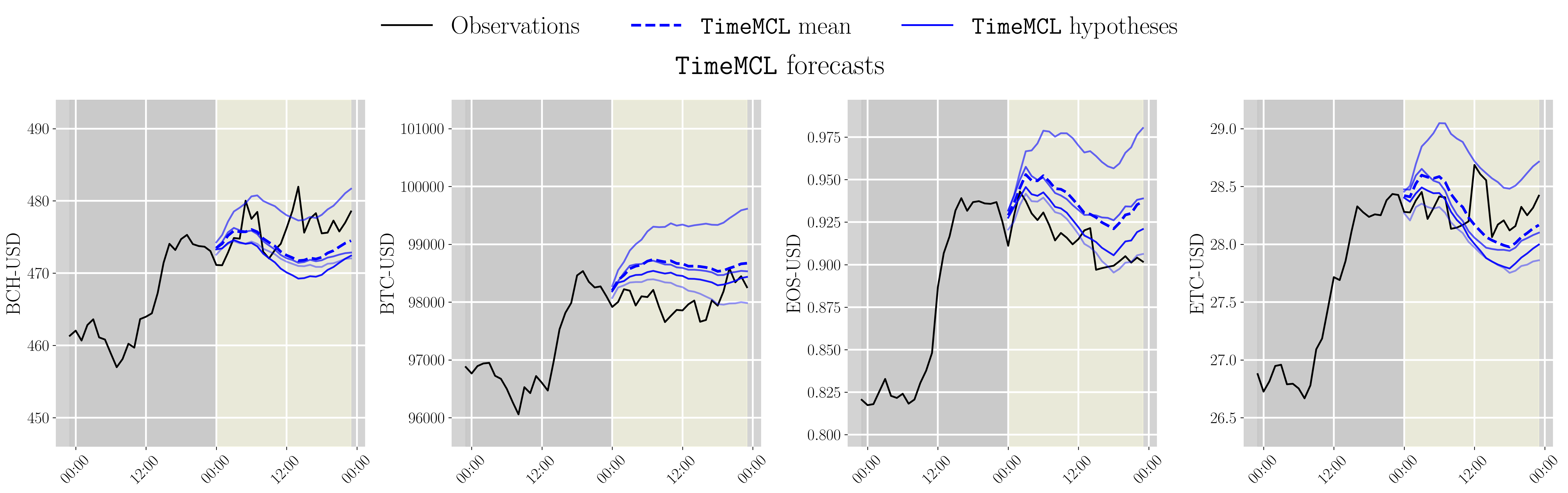}
    \caption{\textbf{Cryptocurrency price forecasting for January 4, 2025 with $K=4$ hypotheses.} Setup in described in Section \ref{sec:financial}. Legend is the same as in Figure \ref{fig:combined_plot_main_paper}. We see that \texttt{TimeMCL} produces at least one hypothesis that closely matches the realized trajectory. We expect each hypothesis to specialize in a distinct market scenario, for example, a trend reversal or a sudden collapse.
    }
    \label{fig:main_paper_crypto_predictions_tmcl}
\end{figure*}

\subsection{Effect of the number of hypotheses}

Table \ref{tab:solar} presents a comparison of performance as a function of the number of hypotheses on the test split of the \textsc{Solar} dataset. See also Table \ref{tab:computation_time_exchange_rate_nips} for an evaluation of the run-time as a function of the number of hypotheses. As expected, the methods generally show improved performance as the number of hypotheses increases, although this also leads to longer run times. However, the performance improvement is not always strictly monotonic with \texttt{TimeMCL}. This suggests that our method still has room for refinement, as some hypotheses may remain slightly underutilized. We suspect that this may be partially due to the choice of scaling by the mean, which could be suboptimal for initializing  \texttt{TimeMCL}. Indeed, as mentioned above, we scale the data by dividing by the mean estimated dimension by dimension, which generally results in a target of constant sign when time series values lie far from the origin, which may be suboptimal when hypotheses are randomly initialized around the origin. To address this, we plan to investigate the impact of different scalers and data pre-processing techniques, such as Z-score normalization or reversible instance normalization \citep{kim2021reversible}, which could promote better use of hypotheses.

\begin{table}
\renewcommand{\arraystretch}{1.1}
    \begin{center}
    \caption{\textbf{Effect of the number of hypotheses on the Distortion Risk for \textsc{Solar}.} Results are averaged over four training seeds. \textbf{\texttt{TimeMCL(R.)}} and \textbf{\texttt{TimeMCL(A.)}} corresponds to the relaxed and annealed variants. See Table \ref{tabapx:solar} for the full results.} 
    \vspace{5pt}
    \label{tab:solar}
    \Large
\resizebox{\columnwidth}{!}{\begin{tabular}{llll||ll}
\toprule
$K$ & \textbf{\texttt{Tactis2}} & \textbf{\texttt{TimeGrad}} & \textbf{\texttt{TempFlow}} & \textbf{\texttt{TimeMCL(R.)}} & \textbf{\texttt{TimeMCL(A.)}} \\
\midrule
1  &              433.76 \scriptsize{$\pm$ 19.02} &  \textbf{398.96 \scriptsize{$\pm$ 32.38}} &  \underline{422.24 \scriptsize{$\pm$ 10.64}} &                                          422.75 \scriptsize{$\pm$ 84.28} &              422.75 \scriptsize{$\pm$ 84.28} \\
2  &              410.23 \scriptsize{$\pm$ 12.98} &  \textbf{380.99 \scriptsize{$\pm$ 31.06}} &              404.89 \scriptsize{$\pm$ 20.53} &  \underline{384.28 \scriptsize{$\pm$ 19.92}} &              418.27 \scriptsize{$\pm$ 95.95} \\
3  &              377.71 \scriptsize{$\pm$ 34.55} &            377.1 \scriptsize{$\pm$ 30.87} &              398.96 \scriptsize{$\pm$ 23.97} &     \textbf{356.42 \scriptsize{$\pm$ 44.11}} &  \underline{358.48 \scriptsize{$\pm$ 20.29}} \\
4  &              376.91 \scriptsize{$\pm$ 34.72} &           375.25 \scriptsize{$\pm$ 32.23} &               386.28 \scriptsize{$\pm$ 18.5} &     \textbf{326.62 \scriptsize{$\pm$ 20.89}} &  \underline{347.45 \scriptsize{$\pm$ 14.59}} \\
5  &  \underline{374.76 \scriptsize{$\pm$ 32.31}} &           374.96 \scriptsize{$\pm$ 32.36} &              385.54 \scriptsize{$\pm$ 19.14} &              376.25 \scriptsize{$\pm$ 43.23} &       \textbf{323.2 \scriptsize{$\pm$ 18.3}} \\
8  &               367.54 \scriptsize{$\pm$ 35.4} &           371.97 \scriptsize{$\pm$ 33.31} &              379.93 \scriptsize{$\pm$ 22.38} &     \textbf{305.63 \scriptsize{$\pm$ 28.24}} &  \underline{347.82 \scriptsize{$\pm$ 28.82}} \\
16 &              358.01 \scriptsize{$\pm$ 38.47} &            360.6 \scriptsize{$\pm$ 34.79} &               371.14 \scriptsize{$\pm$ 18.2} &     \textbf{280.03 \scriptsize{$\pm$ 15.22}} &    \underline{305.5 \scriptsize{$\pm$ 4.05}} \\
\bottomrule
\end{tabular}
}   
\end{center}
\end{table}

\subsection{Financial time series} 
\label{sec:financial}
\texttt{TimeMCL} was evaluated on a corpus of 2 years of hourly cryptocurrency prices
($15$ correlated tickers collected from
\href{https://github.com/ranaroussi/yfinance}{\texttt{YahooFinance}}).
Table~\ref{fig:crypto_dataset_caracteristics} lists the assets, along with
their pairwise correlations and price scales.
Due to the wide variance in price magnitudes across assets, we apply Z-score normalization rather than mean scaling. All models are trained with \(K=4\) hypotheses.
\texttt{TimeMCL} employs the annealed winner-takes-all loss and is compared with
\texttt{Tactis2}, \texttt{TimeGrad}, and \texttt{TempFlow}.
As reported in Table~\ref{tab:results_crypto}, \texttt{TimeMCL} demonstrates strong performance in Distortion, RMSE, and CRPS, while also providing smooth trajectories at a reasonable computation cost (measured in FLOPs). Among the baselines, \texttt{Tactis2} and \texttt{TempFlow} (with a transformer backbone) achieved competitive results, whereas \texttt{TimeGrad}, \texttt{DeepAR}, and the RNN version of \texttt{TempFlow} were less effective. Figure~\ref{fig:main_paper_crypto_predictions_tmcl} shows representative
trajectories of our approach.  At least one \texttt{TimeMCL} hypothesis aligns closely with the
realized path, illustrating that individual hypotheses
can specialize in distinct market regimes (\textit{e.g.,} trend reversals, sudden collapses). Figure~\ref{fig:crypto_price_prediction} contrasts these
predictions with the baselines, highlighting both the smoothness of the
\texttt{TimeMCL} trajectories and their capacity to capture rare events. Implementation details and further analysis can be found in Appendix~\ref{secapx:financial_data}.

\section{Conclusion}
We introduced \texttt{TimeMCL}, a new model for time series forecasting, designed to predict multiple plausible scenarios that can be viewed as an optimal quantization of the future distribution of the process we aim to predict. This model effectively captures different modes of distribution while providing smooth predictions. Moreover, unlike traditional models that rely primarily on the mean and variance of their predictions for interpretation, \texttt{TimeMCL} provides directly interpretable predictions by capturing multiple plausible future scenarios.

\texttt{TimeMCL} could be useful when combined with other backbones, as it complements state-of-the-art approaches with its objective function. In this paper, we implemented the model using a recurrent neural network to model temporal dependencies. Exploring the same approach with different architectures and datasets, such as transformer-based backbones when scaling the dataset size, would be a valuable direction for future research.

\textbf{Limitations.} We found that the choice of scaler can be a limitation in \texttt{TimeMCL}. An inadequate scaler may bias the model toward certain hypotheses. While relaxation techniques help by softening hard winner assignments, we plan to explore advanced normalization methods to further improve \texttt{TimeMCL}’s vanilla setup. Another possible limitation is that \texttt{TimeMCL} requires the number of predictions to be predefined beforehand. Exploring dynamic rearrangements of hypotheses when adding new ones without full retraining is left for future work.

\section*{Acknowledgements}

This work was funded by the French Association for Technological Research (ANRT CIFRE contract 2022-1854) and the LISTEN Laboratory of Télécom Paris. It also benefited from access to the HPC resources of IDRIS (allocation 2024-AD011014345) by GENCI. We are grateful to the reviewers for their insightful comments.

\section*{Impact Statement}

This paper presents work whose goal is to advance the field of Machine Learning. There are many potential societal consequences of our work, none which we feel must be specifically highlighted here.

\bibliography{refs_icml}
\bibliographystyle{icml2025}

\newpage
\appendix
\onecolumn

\section*{Organization of the Appendix}

The Appendix is organized as follows.  
Appendix~\ref{appendixA:theoricalResults} contains the proofs of the theoretical results, establishing that \texttt{TimeMCL} can be interpreted as a functional quantizer. Appendix~\ref{appendix:expr_synth} describes the synthetic data used in the toy example. Appendix~\ref{appendix:expr_reels} details the empirical evaluation on real data, covering the datasets (Appendix~\ref{apx:datasets}), the experimental procedure and evaluation metrics (Appendix~\ref{apx:metrics}), and the baseline models (Appendix~\ref{apx:baselines}).  
Finally, Appendix~\ref{apx:financial_data} presents additional results for the financial time series.

\section{Theoretical results}\label{appendixA:theoricalResults}

Following the main paper notations, we assume each time series lives in $\mathcal{X} = \mathbb{R}^{\dimn}$. We refer to the context and target sequences as $\xPast \in \XPast$ and $\xFuture \in \XFuture$ respectively.

\subsection{Proof of Proposition \ref{prop:opt_quantizer}}

Let us consider the following assumptions.

\begin{assumption}[True risk minimization]
\label{asm:true_risk}
    The batch size is big enough so that the difference between the $\mathcal{L}^{\mathrm{WTA}}$ risk and its empirical version can be neglected.
\end{assumption}

In the Assumption \ref{asm:expressiveness} that follows, we considered the forecaster (or unrolled) network $\mathscr{F}_{\theta}$ as a function $\XPast \rightarrow \XFuture$ that is directly optimized during training, \textit{i.e.,} teacher forcing is disabled. This implies that the same forward computation applies in training and inference. In particular, the model predictions on the target window $\mathcal{X}^{T-t_{0}+1}$ at training time depend only on the past context $\xPast$. We adopt this simplified setup in the toy experiments presented in Section \ref{sec:synthetic} and in Appendix \ref{appendix:expr_synth}.

\begin{assumption}[Expressiveness]
\label{asm:expressiveness}
The family of neural networks considered is expressive enough so that minimizing the expected training risk reduces to minimizing the input-dependent risk for each individual input\footnote{See also Assumption 1 in \citet{perera2024annealed} and Section 3 of \citet{loubes2017prediction}.}. Formally, this means 
\begin{equation}
\underset{\theta \in \Theta}{\mathrm{inf}} \; \mathbb{E}_{x_{1:T}} \left[\ell\left(\mathscr{F}_\theta(\xPast), \xFuture\right)\right] = \int_{\XPast} \underset{z \in \XFuture}{\mathrm{inf}} \; \mathbb{E}_{\xFuture \sim p(\xFuture \mid \xPast)} \left[\ell\left(z, \xFuture\right)\right]\;\mathrm{d}p(\xPast)\;,
\end{equation}
where $\Theta$ is the set of possible neural network parameters, and $\ell$ may be the mean square error loss
$\ell\left(\hat{x}_{t_{0}:T}, \xFuture\right) \triangleq \sum_{t = t_{0}}^{T} \| \hat{x}_t - x_t \|^{2}\;.$
\end{assumption}

\begin{assumption}[Optimality]
\label{asm:optimality}
    The training has converged and $\mathcal{L}^{\mathrm{WTA}}$ has reached a local minima.
\end{assumption}

\begin{proposition}
\label{prop:centroidal_prop_apx}
Under the Assumptions \ref{asm:true_risk}, \ref{asm:expressiveness} and \ref{asm:optimality}, \texttt{TimeMCL} is a conditional stationary quantizer for each sampled window $(\xPast, \xFuture)$, that is 
\begin{equation}
\mathscr{F}_{\theta}^{k}(\xPast) = \mathbb{E}[\xFuture \mid x_{t_0:T} \in \Yk(\xPast)],
\label{eqapx:centroidal}
\end{equation}
where \[
\Yk(\xPast) 
\,=\,
\Bigl\{
\xFuture \in \XFuture \mid
\Lk(\xPast, \xFuture) <
\Lr(\xPast, \xFuture)
\;, \forall r\neq k
\Bigr\},
\]
where accordingly with Assumption \ref{asm:expressiveness}; $\Lk(\xPast, \xFuture) = \sum_{t=t_{0}}^{T} \|\mathscr{F}_{\theta}^{k}(\xPast)_t - x_t \|^{2}$. This makes \texttt{TimeMCL} akin to a conditional and gradient-based version of K-Means over the set of possible future trajectories. 
\end{proposition}

\begin{proof}[Proof]

Assuming we are performing true risk minimization (Assumption \ref{asm:true_risk}), the WTA Loss $\mathcal{L}^{\mathrm{WTA}}$ can be re-written as
\begin{equation}
\label{eqapx:risk}
\mathcal{L}^{\mathrm{WTA}} = \int_{\xPast \in \mathcal{X}^{1:t_{0}}} \sum_{k=1}^{K} \int_{\xFuture \in \Yk(\xPast)} \Lk(\xPast, \xFuture) \; \mathrm{d}p(\xFuture \mid \xPast)  \mathrm{d}p(\xPast)\;,
\end{equation}
due to Chasles relation because the Voronoi tesselation forms a partition of the output space. Under the expressiveness Assumption \ref{asm:expressiveness}, \eqref{eqapx:risk} comes down to optimizing 
\begin{equation}
\mathcal{F}_{\xPast}(z^1,\dots,z^{K}) \triangleq \sum_{k=1}^{K} \int_{\Yk(\xPast)} \mathcal{L}_{z}^k(\xPast, \xFuture) \mathrm{d}p(\xFuture \mid \xPast) \;,
\label{eqapx:functional}
\end{equation}
for each fixed context $\xPast$, where each $z^k \in \XFuture$ and
$$\mathcal{L}_{z}^k(\xPast, \xFuture) \triangleq \sum_{t=t_{0}}^{T} \ell(z^k_t, x_t)\;.$$
To prove that each model $\mathscr{F}_{\theta}^k(\xPast)$ converges to the conditional mean of the future trajectories that fall into its corresponding Voronoi cell, we follow the decoupling strategy of \citet{rupprecht2017learning} and \citet{letzelter24winner}.
Define two sets of variables:
\begin{itemize}
\item \emph{Generators} $\{g^k(\xPast)\}_{k=1}^K$, which induce a partition of the output space $\xFuture$ via the Voronoi diagram:
\[
\Yk(g) \;\triangleq\;
\Bigl\{\xFuture \in \XFuture \mid \|\xFuture - g^k(\xPast)\|^2 
< \|\xFuture - g^r(\xPast)\|^2,\;\forall r \neq k\Bigr\}.
\]
\item \emph{Centroids} $\{z^k(\xPast)\}_{k=1}^K$, which are the points used to compute the intra-cell $L^{2}$ loss.
\end{itemize}
\paragraph{Decoupling generators vs.\ centroids through alternating optimization.} 
In \texttt{TimeMCL} the update is done in a two-step fashion: first, the winners are computed for each point in the batch, and then the latter are updated. This two-step optimization allows us to bypass the non-differentiability of the $\mathrm{min}$ operator in \eqref{eq:wta_loss_bw}. 

Let us \textit{decouple} the two variables and define the following functional to optimize:

\[
\mathcal{F}\bigl(g, z; \xPast\bigr) \triangleq \sum_{k=1}^K
\int_{\Yk(g)} 
\|\xFuture - z^k(\xPast)\|^2 \; p(\xFuture \mid \xPast)\,\mathrm{d}\xFuture\;.
\]

We now consider minimizing $\mathcal{F}(g, z; \xPast)$ with respect to $g$ and $z$ in an alternating manner:

\begin{itemize}
\item \textbf{(a) Fixing the partition ($g$) and optimizing the centroids ($z$).} 
In this case, the gradient update of $z^k$ has direction:
\[
\frac{\partial \mathcal{F}}{\partial\,z^k(\xPast)}(g,z,\xPast)
= 2 \left( z^k(\xPast)\mathrm{Vol}(\Yk(g)) - \int_{\Yk(g)} \xFuture p(\xFuture \mid \xPast) \mathrm{d}\xFuture \right) \;,
\]
where $\mathrm{Vol}(\Yk(g)) \triangleq \int_{\Yk(g)} p(\xFuture \mid \xPast) \mathrm{d}\xFuture$. Thus each $z^k(\xPast)$ is updated in the direction of the cell conditional mean, similarly to the Lloyd algorithm \cite{lloyd1982least}.

\item \textbf{(b) Fixing the centroids ($z$) and optimizing the partition ($g$).}
Conversely, if $z$ is fixed, one can reduce $\mathcal{F}(g, z; \xPast)$ by ensuring that $\Yk(g)$ is indeed the Voronoi cell generated by $z^k(\xPast)$.  Indeed, for any deviation from a strict Voronoi partition, there would exist subsets of $\xFuture$ incorrectly assigned to some cell, and reassigning them to the nearest $z^k(\xPast)$ would lower the overall loss (See also \citet{rupprecht2017learning}, Theorem 1).
\end{itemize}
By alternating between these two steps, any stationary point of $\mathcal{F}(g, z; \xPast)$ must satisfy $$z^k(\xPast) \;=\; g^k(\xPast)\;.$$

Such a configuration is a \emph{centroidal Voronoi tessellation} (see, \textit{e.g.,} \citealp{du1999centroidal}).

Reverting to the original notation of the WTA network,
we identify $g^k(\xPast) = z^k(\xPast) = \mathscr{F}_{\theta}^k(\xPast)$ at optimality,
thus each prediction head coincides with the mean of the distribution restricted to its Voronoi cell:
\[
\mathscr{F}_{\theta}^k(\xPast)
~=~
\mathbb{E}\bigl[\xFuture \;\mid\; \xFuture \in \Yk(\xPast)\bigr]\;.
\]
Consequently, at the global minimum of the WTA objective, we obtain the desired \emph{conditional} centroidal Voronoi tessellation.
\end{proof}

\subsection{Proof of Proposition \ref{prop:opt_quantizer_scores} (Scoring Heads as Unbiased Estimators)}
\label{appendix:proof_scoring_heads}

Once trained, predictions from score heads are performed through the unrolled version of the score heads $\Gamma_{\theta}^k$, which can be viewed as a function \(\Gamma_{\theta}^k : \XPast \rightarrow [0,1]\) when averaging the predicted scores over the sequence. In the following, we assume that the $\Gamma^k_{\theta}$ is directly optimized during training, similarly to the approach used for the prediction heads in Section \ref{appendixA:theoricalResults}. We prove that when the training of \texttt{TimeMCL} with $\beta>0$ has converged globally, the (unrolled) scoring heads 
\(\Gamma_{\theta}^k(\xPast)\) match the conditional probability mass of their respective Voronoi region 
\(\Yk(\xPast)\). This statement generalizes the arguments from \citet{letzelter24winner} to our time-series setup.

\begin{assumption}[Global optimality for both centroids \emph{and} scores]
\label{assump:global_opt}
In addition to having converged to a global minimum of the centroid objective \eqref{eqapx:risk}, the \texttt{TimeMCL} model also reaches a global minimum for its scoring objective
\[
\mathcal{L}^{\mathrm{s}}(\theta) =
\int_{\XPast}
\sum_{k=1}^K 
\int_{\mathcal{X}_k(\xPast)}
\text{BCE}\Bigl(\mathds{1}[\xFuture \in \Yk(\xPast)],\, \Gamma_{\theta}^k(\xPast)\Bigr)\,
p(\xPast,\xFuture)\,\mathrm{d}\xFuture\,\mathrm{d}\xPast\;,
\]
where $\text{BCE}(\cdot,\cdot)$ denotes the binary cross-entropy, and 
$\xPast \mapsto (\Yk(\xPast))_{k=1}^K$ is the optimal Voronoi tessellation 
corresponding to the converged heads $\{\mathscr{F}_\theta^k\}$.
\end{assumption}

\begin{proposition}[Unbiased Estimator of Voronoi Cell Mass]\label{proof:unbiaised_estimator_voronoi_cell} 
\label{prop:unbiased_score_timeseries}
We assume, as in Assumption \ref{asm:true_risk}, that the batch size is big enough so that the difference between the $\mathcal{L}^{\mathrm{s}}$ and its empirical version can be neglected. Under Assumption~ \ref{assump:global_opt}, and assuming perfect expressiveness of the score heads as in Assumption \ref{asm:expressiveness}, for any context $\xPast$ and index $k \in \{1,\dots,K\}$, 
the optimal scoring head satisfies
$$\Gamma_{\theta}^k(\xPast) = \mathbb{P} \left(\xFuture \in \Yk\left( \xPast\right) \mid \xPast \right).
$$
\end{proposition}

\begin{proof}
We adapt the derivation of the unbiased property to our time-series setup. Recall that the optimal scoring objective (see Assumption~\ref{assump:global_opt}) may be written as
\begin{equation}
\label{eq:score_objective}
\min_{\theta}\;
\int_{\XPast}
\sum_{k=1}^K 
\int_{\mathcal{X}_k(\xPast)}
\text{BCE}\Bigl(\,\mathds{1}[\xFuture \in \Yk(\xPast)],\, \Gamma_{\theta}^k(\xPast)\Bigr)\;
p(\xFuture\mid\xPast)\,p(\xPast)\,\mathrm{d}\xFuture\,\mathrm{d}\xPast.
\end{equation}
Under the true risk minimization and expressiveness assumptions, and since the input-dependent risk decomposes over each $k \in \{1,\dots,K\}$, we can focus on a \emph{fixed} $\xPast$ and a single index $k$ in the sum. For that fixed $\xPast$ and $k$, the part of \eqref{eq:score_objective} to optimize is:
\[
\int_{\XFuture} 
\text{BCE}\Bigl(\,\mathds{1}[\xFuture \in \Yk(\xPast)],\, \Gamma_{\theta}^k(\xPast)\Bigr)\;
p(\xFuture\mid\xPast)\,\mathrm{d}\xFuture\;.
\]
Writing out the binary cross-entropy explicitly, we get:
\[
-\int_{\Yk(\xPast)}
\log\bigl(\Gamma_{\theta}^k(\xPast)\bigr)\,p(\xFuture\mid\xPast)\,\mathrm{d}\xFuture
~-\int_{\XFuture \setminus \Yk(\xPast)}
\log\bigl(1 - \Gamma_{\theta}^k(\xPast)\bigr)\,p(\xFuture\mid\xPast)\,\mathrm{d}\xFuture.
\]
Let us denote $m_k(\xPast) = \mathbb{P}\bigl(\xFuture \in \Yk(\xPast) \mid\xPast\bigr)$. Then the above expression becomes
\[
-\,m_k(\xPast)\,\log\bigl(\Gamma_{\theta}^k(\xPast)\bigr)
~-\,(1 - m_k(\xPast))\,\log\bigl(1 - \Gamma_{\theta}^k(\xPast)\bigr)\;.
\]
We recognize this as the binary cross-entropy between the probabilities $\Gamma_{\theta}^k(\xPast)$ and $m_k(\xPast)$. 
The unique global minimum of that scalar binary cross-entropy is attained at
\[
\Gamma_{\theta}^k(\xPast)
~=~
m_k(\xPast)
~=~
\mathbb{P}\bigl(\xFuture \in \Yk\left( \xPast\right) \mid \xPast)\;.
\] 
This equality must hold for each context $\xPast$ and each index $k$, completing the proof.
\end{proof}

\section{Experimental details on the synthetic datasets}\label{appendix:expr_synth}

All toy examples were conducted using a neural network with a three-layer fully connected backbone (\(200\) hidden units, ReLU activation) and \( K \) prediction heads, each generating a trajectory of a given length. Training uses a Winner-Takes-All loss combined with squared error minimization, optimized with Adam using a learning rate of $10^{-3}$, a batch size of $4096$, and $500$ iterations. For each of the setups, we used the Relaxed version of \texttt{TimeMCL} with $\varepsilon = 0.05$.

\paragraph{Brownian Motion.} In order to conduct interpretable experiments and generate synthetic data, we employed Brownian motion. A Brownian motion $\{W_t\}_{t \geq 0}$ is a Gaussian process characterized by a mean $\mathbb{E}[W_t] = 0$ and a covariance function $\mathrm{Cov}(W_t, W_s) = \min(t, s)$ for $t, s \geq 0$. This process possesses several properties of interest \cite{karatzas1998brownian}. In particular, the Markov property, which ensures that the process $\{W_{s + t} - W_{s}\}_{t \geq 0}$ is itself a Brownian motion starting at $0$, independent of the process $\{W_{t} : 0 \leq t \leq s\}$. This implies that to condition a Brownian motion on its trajectory up to time $t$, it suffices to condition on its value $W_t$. The process beyond time $t$ behaves as if it were a new Brownian motion originating at $W_t$. In conditional quantization, we aim to quantize Brownian motion at a given prediction horizon, conditioned on its value at time \(t\). To compute the target quantization, we followed the methodology detailed in \cite{pages2009optimal}, leveraging the Karhunen-Loève decomposition \cite{arcozzi2015karhunen} of the Brownian motion on $t \in [0,1]$:
\begin{equation}
\label{eqapx;karhunen}
W_t = \sum_{n=1}^\infty \sqrt{\lambda_{n}^{W}} \xi_n e_n^W(t)\;,
\end{equation}
where \(\xi_n \sim \mathcal{N}(0, 1)\) are independent standard Gaussian random variables. Here, \(\lambda_{n}^{W} \triangleq \frac{1}{\pi^2 (n - \frac{1}{2})^2}\) are the eigenvalues of the covariance operator of Brownian motion, and \(e_n^W(t) \triangleq \sqrt{2} \sin(\pi (n - \frac{1}{2}) t)\) are the corresponding eigenfunctions \cite{corlay2010nystr}. To quantize Brownian motion, we first truncate its Karhunen-Loève decomposition \eqref{eqapx;karhunen} to \(m\) terms. Then, the coefficients \(\xi_n\) are quantized into \(K_n\) optimal levels \(\{\alpha_1^{(K_n)}, \ldots, \alpha_{K_n}^{(K_n)}\}\), using the quantile function of the normal distribution. All possible combinations of quantized coefficients are generated via the Cartesian product. In this case, for each multi-index $i = (i_1,\dots,i_{m}) \in \prod_{n=1}^{m} [\![1,K_n]\!]$, it is possible to define a quantizer
\[
x_i^{(K_1,\dots,K_m)}(t) \triangleq \sum_{n=1}^{m} \sqrt{\lambda_{n}^{W}} \alpha_{i_n}^{(K_n)} e_n^W(t)\;,
\]
which takes $\prod_{n=1}^{m} K_{n}$ possible values. Following \citet{pages2009optimal}, product quantization with exactly $K$ codevectors requires optimizing two parameters: the truncation parameter $m$ and the quantization level allocation $\{K_n\}_{n=1}^{m}$, with $\prod_{n=1}^{m} K_n = K$. \citet{pages2009optimal} provides a list of optimal tuples for these parameters. In our numerical example, we select $m = 2$ and $(K_1,K_2) = (5,2)$ resulting in a quantizer with $5 \times 2 = 10$ codevectors.

\texttt{TimeMCL} was trained on 250-step sequences randomly drawn from a Brownian motion on \([0,1]\), discretized with 500 steps. At each iteration, a new sample was generated, with the model conditioned on the last observed value and tasked with predicting the next 249 steps. The generated trajectories are smooth and closely match the theoretical optimal ones. The model demonstrates consistent long-term predictions and effectively learns conditional quantization.

\paragraph{Brownian Bridge.} Brownian motion is unrepresentative of most natural processes, as its future evolution is independent of the conditioning time. In contrast, many real-world processes depend on when they are observed. A Brownian bridge, viewed as a Brownian motion conditioned to reach a fixed value at \( T=1 \), better captures such dependencies.
 A continuous process \(\{B_t\}_{t \in [0, T]}\) is a Brownian bridge on the interval \([0, T]\) if and only if it has the same distribution as \(\{W_t - \frac{t}{T} W_T\}_{t \in [0, T]}\), where \(\{W_t\}_{t \in [0, T]}\) is a standard Brownian motion. In our case, \(T = 1\), and the process \(B\) is referred to as a standard Brownian bridge. As we need to conditionally quantize the Brownian bridge, we note that if we pick a random time $t \in [0,1]$, the Brownian bridge conditioned on its value at this point remains a Brownian bridge. To obtain the optimal quantization, the same process as for Brownian motion is used, with adapted eigenvalues and eigenvectors in the Karhunen-Loève decomposition (See \citet{corlay2010nystr}, page 3),
\[
e_n^B(t) \triangleq \sqrt{\frac{2}{T}} \sin\left(\pi n \frac{t}{T}\right), \quad \lambda_n^B \triangleq \frac{T^2}{\pi^2 n^2}\;, \quad n \geq 1\;.
\]

In our numerical example, we select the same values for $m$ and $(K_1,K_2)$ as those chosen for the Brownian motion. \texttt{TimeMCL}, like for Brownian motion, is trained on random samples from a Brownian bridge starting at $B_{0} = 1$ and ending at $B_{1} = 1$. With 500 discretization points, it predicts 250 steps, conditioned on the prediction start time and value. 

\paragraph{Autoregressive model AR(p).} The two first experiments involved Brownian motion, where only the last value matters for prediction, and the Brownian bridge, which depends on the last observed time and value. While these processes assess \texttt{TimeMCL}’s quantization ability, they lack strong context dependence. Real-world applications require modeling context to predict and quantize future trajectories effectively. Consider an autoregressive process
\begin{equation}
    X_t = \sum_{i=1}^{p} \phi_i X_{t-i} + \epsilon_t\;,
\end{equation}
where $(\epsilon_t)_t$ is a white noise with variance $\sigma^2$. We generate an autoregressive sequence where the first five values follow a normal distribution with scale \(\sigma\). The process runs for an initial warm-up period of 100 steps. After this, we predict and quantize the next 250 steps using the last 100 observations, i.e., $\{X_{t_0 - 99}, \dots, X_{t_0}\}$. The five values preceding \(t_0\) form a crucial context—without it, correctly quantizing the future distribution is impossible. \texttt{TimeMCL} is expected to exploit this information for optimal quantization. To the best of our knowledge, no simple analytical solution exists for quantizing an autoregressive process. To approximate an optimal quantization, the process is simulated multiple times from \(t_0\), conditioned on its context, to construct a conditional distribution. The Lloyd algorithm, run with a very large number of simulations ($10^{5}$ sampled trajectories), produces a set of \textit{quantized} trajectories that serve as references for \texttt{TimeMCL}.

\texttt{TimeMCL} is trained on 250-step sequences randomly sampled from 500-step trajectories of the AR(5) process, with a new batch generated at each iteration. To ensure consistency with Brownian bridge and motion experiments, time is normalized by dividing all steps by the total number of steps. We took $\phi_1 = 0.4, \phi_2 = \phi_3 = 0.2, \phi_4 = \phi_5 = 0.1$, and $\sigma = 0.06$.
The trajectories predicted by \texttt{TimeMCL} closely align with the optimal ones found by Lloyd’s algorithm, exhibiting smooth behavior. Various values and orders of the autoregressive process, both stationary and non-stationary, were tested, consistently yielding results aligned with Lloyd’s algorithm.  

\section{Experiments with real data}
\label{appendix:expr_reels}

\subsection{Datasets}
\label{apx:datasets}

We evaluate our approach on six well-established benchmark datasets taken from \texttt{Gluonts} \cite{alexandrov2020gluonts}, each containing strictly positive and bounded real-valued series. See \citet{lai2018modeling} for an extensive description.

\begin{itemize}
\item The \textsc{Solar} dataset \cite{lai2018modeling} consists of hourly aggregated solar power outputs from 137 photovoltaic sites, spanning 7009 time steps. Strong daily seasonality is typically observed due to the day–night cycle.
\item The \textsc{Electricity} dataset \cite{asuncion2007uci} contains hourly power consumption data from 370 clients across 5833 time steps. Demand patterns often exhibit both daily and weekly periodicities, driven by regular human activity and business operations.
\item The \textsc{Exchange} dataset \cite{lai2018modeling} features daily exchange rates for eight different currencies, with 6071 observations per series. Unlike energy or traffic data, exchange rates often lack clear seasonal patterns and are influenced by broader macroeconomic factors.
\item The \textsc{Traffic} dataset \cite{asuncion2007uci} comprises occupancy rates (ranging from 0 to 1) measured hourly by 963 road sensors over 4001 time steps. These series generally display recurrent rush-hour peaks as well as differences between weekdays and weekends.
\item The \textsc{Taxi} dataset consists of traffic time-series data of New York City taxi rides, recorded at 1214 locations every 30 minutes in January 2015 (training set) and January 2016 (test set). We used the preprocessed version from \cite{salinas2019high}.
\item The \textsc{Wikipedia} dataset \cite{gasthaus2019probabilistic} contains daily views of $2000$ Wikipedia pages.
\end{itemize}

Table~\ref{tab:datasets_characteristics} provides an overview of the main characteristics of these six datasets. Note for each dataset, we used the official train/test split. We dedicate 10 times the number of prediction steps for validation, at the end of the training data.  

\begin{table}[h!]
\caption{\textbf{Datasets characteristics.}}
\vspace{4pt}
\centering
\begin{tabular}{lccccc}
\hline
\textsc{Dataset} & \textsc{Dimension} $\dimn$ & \textsc{Domain} $\mathcal{X}$ & \textsc{Freq.} & \textsc{Time Steps} & \textsc{Pred steps}  $T-t_{0}+1$ \\ \hline
\textsc{Solar}           & 137           & $\mathbb{R}^+$      & \textsc{Hour}           & 7,009               & 24                   \\
\textsc{Electricity}      & 370           & $\mathbb{R}^+$      & \textsc{Hour}           & 5,833               & 24                   \\
\textsc{Exchange}         & 8             & $\mathbb{R}^+$      & \textsc{Day}            & 6,071               & 30                   \\
\textsc{Traffic}          & 963           & $(0,1)$      & \textsc{Hour}           & 4,001               & 24                   \\ 
\textsc{Taxi} & 1,214 & $\mathbb{N}$ & \textsc{30-Min} & 1,488 & 24 \\
\textsc{Wikipedia} & 2,000 & $\mathbb{N}$ & \textsc{Day} & 792 & 30 \\ \hline
\end{tabular}
\label{tab:datasets_characteristics}
\end{table}

\subsection{Metrics}
\label{apx:metrics}
In all the following, each model produces $K$ hypotheses $\{\hat{x}^{k}_{\tT}\}_{k}$. Additionally \texttt{Time-MCL} predicts $K$ scores (one per trajectory), that we denote as $\hat{\gamma}_1, \dots, \hat{\gamma}_K \in [0,1]$, with $\sum_{k=1}^{K} \hat{\gamma}_k = 1$, omitting the input $\xPast$ for conciseness. The observed trajectory is denoted as $x_{\tT} \in \mathbb{R}^{\dimn \times(T-t_0+1)}$. We used the \texttt{Gluonts} integrated \texttt{Evaluator} to compute the metrics, and we customized it to include the distortion. For evaluation over the whole test set, we have $N$ input-targets pairs $(x_{\tPast}^{i},x_{\tT}^{i})$, $ i \in \{1, \dots,N\}$. We denote by $\hat{x}^{k,i}_{t}$ the $k$-th hypothesis generated for input $x_{\tPast}^{i}$ at time $t$.

In the following, we define the \textit{summation over dimension} operator as $\mathcal{A}: x \in \mathbb{R}^{\dimn} \mapsto \sum_{d=1}^{\dimn} x^d \in \mathbb{R}$, where $x^d$ denotes the $d$-th dimension of the vector $x$.

\subsubsection{RMSE (Root mean square error)}
\label{metric:rmse}
Let us denote by $\bar{x} \in \mathbb{R}^{\dimn \times(T-t_0+1)}$ the conditional mean estimator of $p(\xFuture \mid \xPast)$ given the probabilistic model we are evaluating. For \texttt{TimeMCL}, we used $\bar{x} = \sum_{k=1}^{K} \hat{\gamma}_k \hat{x}^{k}_{\tT}$ and for the other methods that do not use score heads, $\bar{x} = \frac{1}{K} \sum_{k=1}^{K} \hat{x}^{k}_{\tT}$. RMSE is defined as:
\begin{equation}\label{eq:rmse_formula}
\mathrm{RMSE} \triangleq \sqrt{\frac{1}{N}\sum_{i=1}^{N}\frac{1}{T-t_0+1}\sum_{t=t_0}^{T}\left(\mathcal{A}(x_{t}^{i}) - \mathcal{A}(\bar{x}_{t}^{i})\right)^{2}}\;.
\end{equation}

In \texttt{Gluonts}, we can access this metric under the name \texttt{m\_sum\_rmse}. The prefix \texttt{m\_sum} indicates in \eqref{eq:rmse_formula} that the RMSE is computed after aggregating the target and the prediction dimension by dimension.

\subsubsection{CRPS (Continuous Ranked Probability Score)}\label{appendix:metric_crps}

In the following, we denote by $Q_{X}(q)$ the quantile of order $q$ a real random variable $X$, with $q \in [0,1]$ (when it exists).

Let us define the random variable $X_t^i$ which takes the value $\mathcal{A}(\hat{x}_t^{k, i})$ with probability $\hat{\gamma}_k$, or uniformly with probability $\frac{1}{K}$ when using no score heads, respectively. Let us introduce:
\begin{align*}
\mathscr{L}_{q} \bigl(x_{\tT}^{i},\hat{x}_{\tT}^{i}\bigr)
&\triangleq 
2 \sum_{t=t_0}^{T}
\left\lvert
  \mathcal{A}({x}_{t}^{i})
  - Q_{X_t^i}(q)
\right\rvert
\!\left(
  \mathds{1} (
    \mathcal{A}({x}_{t}^{i})
    \leq
    Q_{X_t^i}(q))
  - q
\right), \quad
\mathscr{T}(x_{t_0:T}) 
\triangleq \sum_{i=1}^{N}\sum_{t=t_0}^{T} 
\left\lvert \mathcal{A}({x}_{t}^{i}) \right\rvert,
\end{align*}
where $\mathscr{L}_{q}$ and $\mathscr{T}$ are referred as the \texttt{quantileLoss} and \texttt{abs\_target\_sum} in \texttt{Gluonts}.

CRPS~\cite{matheson1976scoring} is computed as 
\begin{equation}
\label{eqapx:crps}
\mathrm{CRPS} \triangleq \frac{1}{ |\mathcal{Q} |}\sum_{q \in \mathcal{Q}}\frac{ \sum_{i=1}^{N} \mathscr{L}_{q}\bigl(x_{t_0:T}^{i},\hat{x}_{t_0:T}^{i}\bigr)}{\mathscr{T}(x_{t_0:T})}\;,\end{equation}
where we used $\mathcal{Q} = \{0.05, 0.1 , \dots 0.95\}$. Equation \eqref{eqapx:crps} is referred to as \texttt{m\_sum\_mean\_wQuantileLoss} in the python library, or CRPS-Sum \cite{salinas2019high, ashok2024tactis} in the related literature.

\subsubsection{Distorsion}
\label{metric:distorsion} 

Distortion, also referred to as the \textit{Oracle} or \textit{Quantization} error in the literature \cite{pages2009optimal, lee2016stochastic, perera2024annealed} was defined as
$$D_{2} \triangleq \frac{1}{N}\sum_{i=1}^{N} \min_{k \in \{1,\dots, K\}} \sqrt{\sum_{d=1}^{\dimn}\frac{1}{T-t_0+1}\sum_{t=t_0}^{T}(x_{t}^{i,d} -\hat{x}_{t}^{i,k,d})^{2}}\;.$$
This metric was integrated into the \href{https://ts.gluon.ai/experimental/api/evaluation.html}{\texttt{Evaluator}} class from \texttt{Gluonts}.

\subsubsection{FLOPs (Floating Point Operations)}
\label{secapx:flops}

FLOPs serve as an approximate measure of the model's computational load. For each baseline, we only considered FLOPs evaluation at inference, with a single batch of size $1$ on the \textsc{Exchange} dataset. We used the \texttt{FlopCountAnalysis} function from the \href{https://github.com/facebookresearch/fvcore/blob/main/docs/flop_count.md}{fvcore} library to measure it.

\subsubsection{Inference Run-time} 
\label{secapx:inference_time}

Inference time was computed on a single NVIDIA GeForce RTX 2080 Ti, while making sure it is the only process that runs on the node. It was measured over the whole test set of \textsc{Exchange} (\texttt{make\_evaluation\_predictions} function in \texttt{Gluonts}), using a batch size of $64$. For fair and accurate time comparison with respect to the number of hypotheses, we disable parallel sampling for each baseline (by setting \texttt{num\_parallel\_samples} to $1$).

\subsubsection{Total Variation} 

We quantify the average smoothness of the sampled trajectories using the Total Variation (TV). Given $K$ sampled trajectories, we defined it as
\begin{equation}
\mathrm{TV} \triangleq \frac{1}{N} \sum_{i=1}^{N} \left( \frac{1}{K} \sum_{k=1}^{K} \sum_{t=t_{0}}^{T} \left\|\hat{x}_{t+1}^{k,i}-\hat{x}_{t}^{k,i}\right\|_2 \right)\,.
\label{eqapx:tv}
\end{equation}
For \texttt{TimeMCL}, we sampled the hypotheses in proportion to their scores for Total Variation computation instead of uniformly as in \eqref{eqapx:tv} for the other baselines. The smoother the predictions of a probabilistic estimator, the smaller its total variation.

\subsection{Experimental details}\label{apx:baselines}

\begin{table}
    \begin{center}
    \caption{\textbf{Distortion comparison with 8 Hypotheses.} Results are averaged over four training seeds. \textbf{\texttt{ETS}}, \textbf{\texttt{Trf.TempFlow}} and \textbf{\texttt{Tactis2}}, columns are in \textcolor{gray}{gray} because they don't share the same backbone as the other baselines. Best scores are in \textbf{bold}, and second-best are \underline{underlined}.}
    \label{tab:dist_8hyp}
    \resizebox{\columnwidth}{!}{
    \begin{tabular}{lcccccc||cc}
\toprule
{}  & \textcolor{gray}{\textbf{\texttt{ETS}}} & \textcolor{gray}{\textbf{\texttt{Trf.TempFlow}}} & \textcolor{gray}{\textbf{\texttt{Tactis2}}} &      \textbf{\texttt{TimeGrad}}  &  \textbf{\texttt{DeepAR}}  &  \textbf{\texttt{TempFlow}}  &  \textbf{\texttt{TimeMCL(R.)}}  &   \textbf{\texttt{TimeMCL(A.)}} \\
\midrule
\textsc{Elec.}    & \textcolor{gray}{15132 $\pm$ 1662} & \textcolor{gray}{14160 $\pm$ 3370} & \textcolor{gray}{12557 $\pm$ 2405} &        \textbf{10172 $\pm$ 680}  &         144832 $\pm$ 3081  &             15822 $\pm$ 531  &                 12155 $\pm$ 414  &      \underline{11509 $\pm$ 703} \\
\textsc{Exch.}    & \textcolor{gray}{0.057 $\pm$ 0.002} & \textcolor{gray}{0.066 $\pm$ 0.012} & \textcolor{gray}{\textbf{0.031 $\pm$ 0.002}} &               0.041 $\pm$ 0.006  &         0.062 $\pm$ 0.001  &           0.051 $\pm$ 0.007  &   \underline{0.039 $\pm$ 0.001}  &                0.042 $\pm$ 0.005 \\
\textsc{Solar}    & \textcolor{gray}{654.48 $\pm$ 3.45} & \textcolor{gray}{387.15 $\pm$ 18.57} & \textcolor{gray}{367.54 $\pm$ 35.4} &              371.97 $\pm$ 33.31  &        770.48 $\pm$ 19.91  &          379.93 $\pm$ 22.38  &     \textbf{305.63 $\pm$ 28.24}  &   \underline{347.82 $\pm$ 28.82} \\
\textsc{Traffic}  & \textcolor{gray}{2.65 $\pm$ 0.0} & \textcolor{gray}{1.28 $\pm$ 0.02} & \textcolor{gray}{0.85 $\pm$ 0.04} &                 0.79 $\pm$ 0.05  &           2.14 $\pm$ 0.06  &             1.23 $\pm$ 0.04  &        \textbf{0.69 $\pm$ 0.02}  &      \underline{0.71 $\pm$ 0.01} \\
\textsc{Taxi}     & \textcolor{gray}{589.15 $\pm$ 0.75} & \textcolor{gray}{284.25 $\pm$ 13.94} & \textcolor{gray}{245.09 $\pm$ 8.79} &   \underline{211.71 $\pm$ 4.02}  &        429.01 $\pm$ 14.88  &          272.84 $\pm$ 12.14  &      \textbf{187.04 $\pm$ 2.45}  &               256.16 $\pm$ 24.95 \\
\textsc{Wiki}     & \textcolor{gray}{722166 $\pm$ 5248} & \textcolor{gray}{528546 $\pm$ 11657} & \textcolor{gray}{\textbf{255328 $\pm$ 3386}} &               261463 $\pm$ 1047  &         371035 $\pm$ 6578  &           388511 $\pm$ 2472  &              271878 $\pm$ 17358  &    \underline{258468 $\pm$ 3108} \\
\bottomrule
\end{tabular}
}
    \end{center}
    \end{table}

\subsubsection{Architectures, training, and inference}

\noindent
\textbf{Recurrent Neural Network Backbone.} \texttt{Time-MCL}, \texttt{TempFlow}, \texttt{TimeGrad}, and \texttt{DeepAR} share the same RNN backbone, which takes as input a concatenation of feature types. Let $B$ be the batch size, $T$ the time series length, and $D_{\text{target}}$ the target dimension. The input includes $(i)$ lagged target values, normalized and shaped as $B \times T \times (n_{\text{lags}} \times D_{\text{target}})$, and $(ii)$ Fourier features of shape $B \times T \times (2 \times N_f)$. The combined input to the RNN thus has shape $B \times T \times \left((n_{\text{lags}} \times D_{\text{target}}) + (2 \times N_f)\right)$. Our RNN implementation follows \citet{rasul2021autoregressive} and consists of $L = 2$ layers of Long Short-Term Memory (LSTM) cells, each containing $C = 40$ hidden units. The term $N_f$, which is associated with Fourier transforms, will be further detailed in the following sections.

\textbf{Sampling and dataloader} The time series is first segmented into past and future windows based on predefined lengths and lead time—sampling slicing points randomly during training and using the last timestamp during prediction (\texttt{InstanceSlicer}). After transformation, the dataloader assembles inputs used across all models. It separates past and future targets and adds several structured features: $(i)$ a \textit{Target Dimension Indicator} ($B \times D_{\text{target}}$) uniquely labels each target dimension; $(ii)$ an \textit{Observed Values Indicator} replaces missing target entries with dummies and flags them as observed (1) or missing (0); $(iii)$ a \textit{Padding Mask} marks synthetic padding to ensure uniform sequence lengths, preventing them from influencing training. Additionally, temporal features of shape $B \times T_{\text{pred}} \times 2N_f$ encode periodic patterns using sine and cosine projections, where $N_f$ denotes the number of unique values for a given frequency (\textit{e.g.,} $N_f = 12$ for months).

\textbf{Transformer Backbone.} For the transformer-based version of TempFlow (\texttt{Trf.TempFlow}) and for \texttt{Tactis2}, we employed the same architectures as in their respective original implementations \cite{rasul2021multivariate, ashok2024tactis}.

\begin{table}
    \begin{center}
    \caption{\textbf{RMSE ($\downarrow$) comparison for $K = 16$ hypotheses.}}
    \label{tab:rmse_sum}
    \resizebox{\columnwidth}{!}{
    \begin{tabular}{lcccccc||cc}
\toprule
{}  & \textcolor{gray}{\textbf{\texttt{ETS}}} & \textcolor{gray}{\textbf{\texttt{Trf.TempFlow}}} & \textcolor{gray}{\textbf{\texttt{Tactis2}}} &         \textbf{\texttt{TimeGrad}}  &  \textbf{\texttt{DeepAR}}  &  \textbf{\texttt{TempFlow}}  &  \textbf{\texttt{TimeMCL (R.)}}  &  \textbf{\texttt{TimeMCL (A.)}} \\
\midrule
\textsc{Elec.}    & \textcolor{gray}{25771 $\pm$ 1117} & \textcolor{gray}{20559 $\pm$ 11011} & \textcolor{gray}{19342 $\pm$ 5496} &        \textbf{11597 $\pm$ 931}  &         145977 $\pm$ 3564  &            23373 $\pm$ 2614  &   \underline{18571 $\pm$ 3063}  &               19154 $\pm$ 2657 \\
\textsc{Exch.}    & \textcolor{gray}{0.089 $\pm$ 0.01} & \textcolor{gray}{0.147 $\pm$ 0.043} & \textcolor{gray}{\textbf{0.062 $\pm$ 0.007}} &   \underline{0.077 $\pm$ 0.027}  &         0.086 $\pm$ 0.002  &           0.095 $\pm$ 0.021  &              0.084 $\pm$ 0.002  &              0.103 $\pm$ 0.021 \\
\textsc{Solar}    & \textcolor{gray}{3661.86 $\pm$ 28.87} & \textcolor{gray}{4121.74 $\pm$ 243.06} & \textcolor{gray}{\underline{3531.02 $\pm$ 439.59}} &   \textbf{3244.92 $\pm$ 247.16}  &      7199.09 $\pm$ 107.71  &        3801.72 $\pm$ 122.09  &           3639.23 $\pm$ 606.34  &           3896.48 $\pm$ 468.71 \\
\textsc{Traffic}  & \textcolor{gray}{15.24 $\pm$ 0.09} & \textcolor{gray}{27.57 $\pm$ 0.22} & \textcolor{gray}{11.54 $\pm$ 1.29} &        \textbf{4.44 $\pm$ 1.07}  &          39.15 $\pm$ 1.48  &            26.77 $\pm$ 1.29  &    \underline{5.48 $\pm$ 0.71}  &                 5.73 $\pm$ 0.7 \\
\textsc{Taxi}     & \textcolor{gray}{9779.81 $\pm$ 13.42} & \textcolor{gray}{4223.88 $\pm$ 561.11} & \textcolor{gray}{\underline{2662.52 $\pm$ 58.79}} &   \textbf{2357.85 $\pm$ 133.05}  &      7323.32 $\pm$ 245.35  &        4436.39 $\pm$ 581.06  &          6855.32 $\pm$ 3456.01  &          8549.15 $\pm$ 2691.98 \\
\textsc{Wiki}     & \textcolor{gray}{765663 $\pm$ 39186} & \textcolor{gray}{1152828 $\pm$ 171868} & \textcolor{gray}{\underline{564176 $\pm$ 115593}} &     \textbf{505869 $\pm$ 60432}  &     4469007 $\pm$ 1015212  &         864607 $\pm$ 105223  &           1210500 $\pm$ 412274  &           1264339 $\pm$ 166347 \\
\bottomrule
\end{tabular}
}
    \end{center}
    \end{table}

\begin{table}
    \begin{center}
    \caption{\textbf{CRPS-Sum ($\downarrow$) comparison for $K = 16$ hypotheses.} Results averaged over four training seeds.}
    \label{tab:crps_sum}
    \resizebox{\columnwidth}{!}{
    \begin{tabular}{lcccccc||cc}
\toprule
{}  & \textcolor{gray}{\textbf{\texttt{ETS}}} & \textcolor{gray}{\textbf{\texttt{Trf.TempFlow}}} & \textcolor{gray}{\textbf{\texttt{Tactis2}}} &     \textbf{\texttt{TimeGrad}}  &  \textbf{\texttt{DeepAR}}  &  \textbf{\texttt{TempFlow}}  &  \textbf{\texttt{TimeMCL (R.)}}  &    \textbf{\texttt{TimeMCL (A.)}} \\
\midrule
\textsc{Elec.}    & \textcolor{gray}{0.0601 $\pm$ 0.0045} & \textcolor{gray}{0.0434 $\pm$ 0.0214} & \textcolor{gray}{\underline{0.0409 $\pm$ 0.0105}} &      \textbf{0.0235 $\pm$ 0.0018}  &       0.4234 $\pm$ 0.0101  &         0.0503 $\pm$ 0.0038  &               0.0488 $\pm$ 0.0074  &              0.0481 $\pm$ 0.0086 \\
\textsc{Exch.}    & \textcolor{gray}{0.0079 $\pm$ 0.001} & \textcolor{gray}{0.0138 $\pm$ 0.0039} & \textcolor{gray}{\textbf{0.0066 $\pm$ 0.0014}} &   \underline{0.0074 $\pm$ 0.0029}  &       0.0081 $\pm$ 0.0002  &         0.0091 $\pm$ 0.0019  &               0.0103 $\pm$ 0.0002  &              0.0131 $\pm$ 0.0031 \\
\textsc{Solar}    & \textcolor{gray}{0.509 $\pm$ 0.0041} & \textcolor{gray}{0.4528 $\pm$ 0.0271} & \textcolor{gray}{0.3919 $\pm$ 0.0378} &      \textbf{0.3789 $\pm$ 0.0356}  &        1.028 $\pm$ 0.0225  &         0.4352 $\pm$ 0.0215  &   \underline{0.3831 $\pm$ 0.0658}  &              0.3966 $\pm$ 0.0079 \\
\textsc{Traffic}  & \textcolor{gray}{0.2007 $\pm$ 0.0014} & \textcolor{gray}{0.394 $\pm$ 0.0092} & \textcolor{gray}{0.1339 $\pm$ 0.0166} &      \textbf{0.0561 $\pm$ 0.0137}  &       0.5257 $\pm$ 0.0205  &         0.3833 $\pm$ 0.0271  &                0.067 $\pm$ 0.0102  &   \underline{0.0653 $\pm$ 0.009} \\
\textsc{Taxi}     & \textcolor{gray}{0.8948 $\pm$ 0.0017} & \textcolor{gray}{0.288 $\pm$ 0.0467} & \textcolor{gray}{\underline{0.1756 $\pm$ 0.0056}} &      \textbf{0.1449 $\pm$ 0.0118}  &       0.5223 $\pm$ 0.0239  &         0.3106 $\pm$ 0.0512  &               0.4438 $\pm$ 0.1965  &              0.5803 $\pm$ 0.2206 \\
\textsc{Wiki}     & \textcolor{gray}{0.0789 $\pm$ 0.0041} & \textcolor{gray}{0.1309 $\pm$ 0.0316} & \textcolor{gray}{\underline{0.0649 $\pm$ 0.019}} &      \textbf{0.0567 $\pm$ 0.0084}  &       0.6588 $\pm$ 0.2225  &         0.0905 $\pm$ 0.0051  &                0.142 $\pm$ 0.0624  &              0.1479 $\pm$ 0.0392 \\
\bottomrule
\end{tabular}
}
    \end{center}
    \end{table}

\textbf{Optimization} Training is conducted using the Adam optimizer with a learning rate of $10^{-3}$, and following an `LR on plateau' scheduler (See \href{https://pytorch.org/docs/stable/generated/torch.optim.lr_scheduler.ReduceLROnPlateau.html}{Pytorch documentation}), a weight decay of $10^{-8}$ and $200$ training epochs. Additionally, a separate validation split is used with a temporal size equal to $10$ times the prediction length. When training \texttt{TimeMCL}, the WTA loss (and its variants) was divided by the prediction length $T-t_{0}+1$. All models are trained using gradient norm clipping with a threshold of 10. Training and evaluation are performed in single-precision (float32) with PyTorch.

\textbf{Inference.} We used the official experimental protocol for evaluation in this benchmark (\textit{e.g.,} \cite{rasul2021autoregressive}). The official test dataset is divided into multiple non-overlapping subsets, each containing sufficient points for both context and prediction lengths, allowing comprehensive assessment across different temporal segments. To compute the \texttt{TimeMCL} metrics while respecting the probabilities given by the scores, we applied a simple trick: after a single forward pass, we obtain $K$ hypotheses along with their associated probabilities (derived from the scores). To ensure that the hypotheses contribute to the metrics in proportion to their assigned probabilities, we performed resampling with replacement from these $K$ hypotheses based on their respective probabilities before computing metrics.

\subsubsection{Baselines}

\textbf{\texttt{DeepAR}.} \cite{salinas2020deepar}. The output distribution as a multivariate normal distribution with mean vector \(\mu \in \mathbb{R}^D\), a low-rank covariance factor \(L \in \mathbb{R}^{D \times r}\), and a diagonal covariance vector \(\sigma \in \mathbb{R}^D\), where the covariance matrix $\Sigma$ is defined as \(\Sigma = LL^\top + \text{diag}(\sigma)\), thereby ensuring efficient parameterization and capturing primary dependencies among the target dimensions. In all the experiments, we set the rank $r$ to $1$.

\textbf{\texttt{TimeGrad}.} \cite{rasul2021autoregressive}.  For the diffusion process, as in \citet{rasul2021autoregressive} we employ $100$ diffusion steps with a linear variance schedule, transitioning from $\beta_1 = 1 \times 10^{-4}$ to $\beta_{100} = 0.1$. The loss function utilized is the $L^{2}$ loss. The residual architecture is integrated within the diffusion component to facilitate effective training and capture complex temporal dependencies. This residual network consists of $8$ residual layers, each with $8$ residual channels. The dilation cycle length is set to $2$, which controls the expansion of the receptive field without increasing the number of parameters. 

\textbf{\texttt{TempFlow}} \citet{rasul2021multivariate}. TempFlow applies a conditional normalizing flow \cite{papamakarios2021normalizing} that is invertible and has a tractable Jacobian, yielding an exact likelihood. Batch normalization after each coupling layer and per-series mean scaling further stabilizes optimization. We adopt the original \texttt{TempFlow} configuration from \citet{rasul2021multivariate}: a two-layer LSTM conditioner with $40$ cells and dropout $0.1$, fed with a $100$-step context, and a Real-valued Non-Volume Preserving (RealNVP) \cite{dinh2017density} head composed of $3$ coupling blocks whose scale/shift networks have $2$ hidden layers with $100$ hidden units. The transformers version of TempFlow \texttt{Trf.TempFlow} uses $8$ heads with $3$ encoder and $3$ decoding layers, with an embedding dimension of $32$, and a feedforward dimension of $128$ with a dropout rate of $0.1$ and Gelu \cite{hendrycks2016gaussian} activation layers. 

\textbf{\texttt{Tactis2}} \cite{ashok2024tactis}. Unlike the RNN-based \texttt{TimeGrad}, \texttt{TempFlow}, and \texttt{Time-MCL}, \texttt{Tactis2} encodes each time step as a token that includes the value, a binary missingness flag, static covariates, and a positional code. Two Transformer encoders process these tokens to produce separate embeddings: one for the marginal distributions and another for the dependence structure. A hyper-network transforms the marginal embedding into the parameters of a Deep-Sigmoidal Flow, which maps the observation to its probability integral transform. An \emph{attentional copula} conditioned on the dependence embedding captures cross-series interactions. The training follows a two-stage curriculum: first, learning the marginals associated with each dimension; then freezing them and fitting the copula. We use the default hyperparameters reported by \citet{ashok2024tactis} for \texttt{fred-md} \cite{godahewa2monash}, detailed in this \href{https://github.dev/servicenow/tactis}{notebook} from the official code. Our setup uses $5$-dimensional time-series embeddings. Both the Marginal CDF Encoder and the Attentional Copulas Encoder have two layers with one attention head (dimension $16$, no dropout). The Transformer Decoder has one attention layer (dimension $8$) with $3$ heads, a $2$-layer MLP (hidden dimension $48$), and $20$ histogram bins. The Deep Sigmoidal Flow marginals \cite{huang2018neural} consist of $2$ flow layers of dimension $8$ and a $2$-layer MLP (hidden dimension $48$). The training comprises $20$ epochs in phase 1 and $180$ epochs in phase 2. Each time series is standardized by subtracting its training-set mean and dividing by its training-set standard deviation, ensuring zero mean and unit variance before model ingestion. Optimization parameters are kept the same as in the other baselines.

\textbf{\texttt{ETS}}. Exponential smoothing \cite{hyndman2008forecasting} is a non-neural model. Holt-Winters Exponential Smoothing was applied with an additive trend and an additive seasonality, assuming a seasonal period of $24$. Model parameters were estimated through automatic optimization using the \texttt{statsmodels} library.

\textbf{\texttt{TimeMCL\,(Relaxed)}} The Relaxed-WTA Loss \cite{rupprecht2017learning} was computed for each pair $(\xPast,\xFuture)$ as \begin{equation}
\label{eq:rwta_loss}
(1-\varepsilon)\; \mathcal{L}_{\theta}^{k^{\star}}(\xPast,\xFuture) + \frac{\varepsilon}{K-1} \sum_{\substack{s=1,\,s \neq k^{\star}}}^{K} \mathcal{L}_{\theta}^{s}(\xPast,\xFuture)\;,\end{equation} with $\varepsilon = 0.1$. We found that this provides a good trade-off for handling under-trained hypotheses without deteriorating the distortion performance. However, we did not specifically tune this value for each dataset.

\textbf{\texttt{TimeMCL\,(Annealed)}} The annealed MCL (aMCL) Loss \cite{perera2024annealed} was computed with an exponential temperature scheduler; $T(t) = T_{0} \rho^{t}$, where $t$ is the number of the training epoch, $T_{0}$ is the initial temperature, and $\rho$ is the decay factor. At each temperature $T$, the aMCL loss is computed for each pair $(\xPast,\xFuture)$ as 
\begin{equation}
\label{eq:amcl_loss}
\sum_{k=1}^{K} q_k(T) \Lk(\xPast,\xFuture)\;,\end{equation} where the coefficients 
\begin{equation}q_{k}(T) \triangleq \frac{1}{Z(\xPast,\xFuture;T)} \exp \left(-\frac{\Lk(\xPast,\xFuture)}{T}\right), \;\;\;  {Z(\xPast,\xFuture;T)} \triangleq \sum_{s = 1}^{K} \exp \left(-\frac{\ell(\Ls(\xPast,\xFuture)}{T} \right),\label{eqapx:awta_assignation}
    \end{equation}

are detached from the computational graph. In our experiments with aMCL, we set $\rho = 0.95$ and $T_{0} = 10$, and we set as limit temperature $T_{\mathrm{lim}} = 5 \times 10^{-4}$ before switching back to the vanilla WTA mode.

\begin{table}
    \begin{center}
    \caption{\textbf{Distorsion ($\downarrow$) Comparison for Solar Dataset.} Results averaged over four training seeds. In this table, the distortion is computed with a variable number of hypotheses $K$ for each baseline, as in Table \ref{tab:solar} of the main paper.}
    \vspace{2pt}
    \label{tabapx:solar}
    \resizebox{\columnwidth}{!}{
    \begin{tabular}{lcccccc||cc}
\toprule
{$K$} & \textcolor{gray}{\textbf{\texttt{ETS}}} & \textcolor{gray}{\textbf{\texttt{Trf.TempFlow}}} &       \textcolor{gray}{\textbf{\texttt{Tactis2}}} &      \textbf{\texttt{TimeGrad}} & \textbf{\texttt{DeepAR}} & \textbf{\texttt{TempFlow}} & \textbf{\texttt{TimeMCL(R.)}} &  \textbf{\texttt{TimeMCL(A.)}} \\
\midrule
1   & \textcolor{gray}{685.77 $\pm$ 23.08} & \textcolor{gray}{468.84 $\pm$ 32.54} & \textcolor{gray}{433.76 $\pm$ 19.02} &   \textbf{398.96 $\pm$ 32.38}  &        838.14 $\pm$ 26.13  &   \underline{422.24 $\pm$ 10.64}  &                             422.75 $\pm$ 84.28  &               422.75 $\pm$ 84.28 \\
2   & \textcolor{gray}{678.49 $\pm$ 13.29} & \textcolor{gray}{424.7 $\pm$ 14.47} & \textcolor{gray}{410.23 $\pm$ 12.98} &   \textbf{380.99 $\pm$ 31.06}  &        817.75 $\pm$ 15.28  &               404.89 $\pm$ 20.53  &   \underline{384.28 $\pm$ 19.92}  &               418.27 $\pm$ 95.95 \\
3   & \textcolor{gray}{678.94 $\pm$ 10.6} & \textcolor{gray}{409.97 $\pm$ 20.37} & \textcolor{gray}{377.71 $\pm$ 34.55} &             377.1 $\pm$ 30.87  &        794.45 $\pm$ 19.39  &               398.96 $\pm$ 23.97  &      \textbf{356.42 $\pm$ 44.11}  &   \underline{358.48 $\pm$ 20.29} \\
4   & \textcolor{gray}{671.97 $\pm$ 16.52} & \textcolor{gray}{402.49 $\pm$ 20.51} & \textcolor{gray}{376.91 $\pm$ 34.72} &            375.25 $\pm$ 32.23  &        789.66 $\pm$ 18.97  &                386.28 $\pm$ 18.5  &      \textbf{326.62 $\pm$ 20.89}  &   \underline{347.45 $\pm$ 14.59} \\
5   & \textcolor{gray}{666.44 $\pm$ 6.8} & \textcolor{gray}{400.63 $\pm$ 18.4} & \textcolor{gray}{\underline{374.76 $\pm$ 32.31}} &            374.96 $\pm$ 32.36  &        776.94 $\pm$ 17.93  &               385.54 $\pm$ 19.14  &               376.25 $\pm$ 43.23  &        \textbf{323.2 $\pm$ 18.3} \\
8   & \textcolor{gray}{654.48 $\pm$ 3.45} & \textcolor{gray}{387.15 $\pm$ 18.57} & \textcolor{gray}{367.54 $\pm$ 35.4} &            371.97 $\pm$ 33.31  &        770.48 $\pm$ 19.91  &               379.93 $\pm$ 22.38  &      \textbf{305.63 $\pm$ 28.24}  &   \underline{347.82 $\pm$ 28.82} \\
16  & \textcolor{gray}{641.32 $\pm$ 3.57} & \textcolor{gray}{374.69 $\pm$ 20.97} & \textcolor{gray}{358.01 $\pm$ 38.47} &             360.6 $\pm$ 34.79  &        748.68 $\pm$ 18.92  &                371.14 $\pm$ 18.2  &      \textbf{280.03 $\pm$ 15.22}  &     \underline{305.5 $\pm$ 4.05} \\
\bottomrule
\end{tabular}
}
    \end{center}
    \end{table}

\begin{table}
    \begin{center}
    \caption{\textbf{Inference time (in seconds) for $\textsc{Exchange}$ dataset.} Results averaged over $15$ random seeds. For accurate time comparison with respect to $K$, we disable parallel computation of the samples for each baseline. \texttt{ETS}, which doesn't require neural networks is also included for completeness (in gray). See Appendix \ref{secapx:inference_time} for details.}
    \vspace{3pt}
    \label{tab:computation_time_exchange_rate_nips}  
    \resizebox{0.9\columnwidth}{!}{
    \begin{tabular}{ccccccc||c}
\toprule
$K$ & \textcolor{gray}{\textbf{\texttt{ETS}}} & \textbf{\texttt{Trf.TempFlow}} & \textbf{\texttt{Tactis2}} & \textbf{\texttt{TimeGrad}} & \textbf{\texttt{DeepAR}} & \textbf{\texttt{TempFlow}} & \textbf{\texttt{TimeMCL}} \\
\midrule
1 & \textcolor{gray}{0.06 $\pm$ 0.01} & 0.19 $\pm$ 0.02 & 0.56 $\pm$ 0.04 & 14.92 $\pm$ 0.23 & \underline{0.09 $\pm$ 0.01} & 0.13 $\pm$ 0.01 & \textbf{0.08 $\pm$ 0.01} \\
2 & \textcolor{gray}{0.06 $\pm$ 0.00} & 0.34 $\pm$ 0.01 & 1.10 $\pm$ 0.05 & 30.04 $\pm$ 0.49 & \underline{0.13 $\pm$ 0.01} & 0.22 $\pm$ 0.01 & \textbf{0.11 $\pm$ 0.01} \\
3 & \textcolor{gray}{0.06 $\pm$ 0.00} & 0.48 $\pm$ 0.02 & 1.63 $\pm$ 0.07 & 45.10 $\pm$ 0.50 & \underline{0.18 $\pm$ 0.01} & 0.30 $\pm$ 0.02 & \textbf{0.14 $\pm$ 0.01} \\
4 & \textcolor{gray}{0.06 $\pm$ 0.00} & 0.63 $\pm$ 0.02 & 2.18 $\pm$ 0.08 & 60.07 $\pm$ 0.97 & \underline{0.22 $\pm$ 0.01} & 0.39 $\pm$ 0.01 & \textbf{0.18 $\pm$ 0.01} \\
5 & \textcolor{gray}{0.07 $\pm$ 0.00} & 0.79 $\pm$ 0.04 & 2.75 $\pm$ 0.20 & 75.69 $\pm$ 0.86 & \underline{0.26 $\pm$ 0.01} & 0.49 $\pm$ 0.05 & \textbf{0.23 $\pm$ 0.02} \\
8 & \textcolor{gray}{0.06 $\pm$ 0.00} & 1.22 $\pm$ 0.05 & 4.32 $\pm$ 0.11 & 119.67 $\pm$ 0.97 & \textbf{0.38 $\pm$ 0.02} & 0.73 $\pm$ 0.03 & \underline{0.39 $\pm$ 0.01} \\
16 & \textcolor{gray}{0.06 $\pm$ 0.00} & 2.47 $\pm$ 0.23 & 8.69 $\pm$ 0.36 & 241.57 $\pm$ 2.24 & \textbf{0.70 $\pm$ 0.04} & 1.39 $\pm$ 0.03 & \underline{1.12 $\pm$ 0.04} \\
\bottomrule
\end{tabular}

    }
    \end{center}
    \end{table}

\begin{table}
    \begin{center}
    \caption{\textbf{Total Variation ($\downarrow$) comparison for $K = 16$ hypotheses.} \textbf{\texttt{ETS}}, \textbf{\texttt{Trf.TempFlow}} and \textbf{\texttt{Tactis2}} columns are in \textcolor{gray}{gray} because they don't share the same backbone as the other baseline. Best methods are \textbf{bold}, second best are \underline{underlined}. Results averaged over four training seeds. Total Variation quantifies the average smoothness of the predicted trajectories (lower is more smooth).}
    \label{tabapx:TV Score}
    \resizebox{\columnwidth}{!}{\begin{tabular}{lcccccc||cc}
\toprule
  {}  & \textcolor{gray}{\textbf{\texttt{ETS}}} & \textcolor{gray}{\textbf{\texttt{Trf.TempFlow}}} & \textcolor{gray}{\textbf{\texttt{Tactis2}}} &  \textbf{\texttt{TimeGrad}}  &  \textbf{\texttt{DeepAR}}  &  \textbf{\texttt{TempFlow}}  &     \textbf{\texttt{TimeMCL(R.)}}  &     \textbf{\texttt{TimeMCL(A.)}} \\
\midrule
\textsc{Elec.}    & \textcolor{gray}{315611 $\pm$ 1858} & \textcolor{gray}{451750 $\pm$ 61031} & \textcolor{gray}{284232 $\pm$ 23269} &          372443 $\pm$ 18499  &       4584308 $\pm$ 97542  &          434097 $\pm$ 34623  &       \textbf{220375 $\pm$ 33961}  &    \underline{245908 $\pm$ 30505} \\
\textsc{Exch.}    & \textcolor{gray}{0.426 $\pm$ 0.0052} & \textcolor{gray}{0.633 $\pm$ 0.0764} & \textcolor{gray}{0.237 $\pm$ 0.0355} &          0.606 $\pm$ 0.0208  &        2.075 $\pm$ 0.0434  &          1.104 $\pm$ 0.1582  &       \textbf{0.031 $\pm$ 0.0075}  &    \underline{0.042 $\pm$ 0.0141} \\
\textsc{Solar}    & \textcolor{gray}{5899 $\pm$ 17} & \textcolor{gray}{3924 $\pm$ 381} & \textcolor{gray}{4421 $\pm$ 828} &             5383 $\pm$ 2173  &           10163 $\pm$ 229  &              3653 $\pm$ 488  &        \underline{3391 $\pm$ 947}  &           \textbf{2195 $\pm$ 297} \\
\textsc{Traffic}  & \textcolor{gray}{18.782 $\pm$ 0.016} & \textcolor{gray}{20.56 $\pm$ 0.932} & \textcolor{gray}{11.063 $\pm$ 0.874} &          12.991 $\pm$ 1.661  &         69.603 $\pm$ 2.61  &          18.114 $\pm$ 0.512  &      \underline{5.766 $\pm$ 0.19}  &        \textbf{5.714 $\pm$ 0.283} \\
\textsc{Taxi}     & \textcolor{gray}{4385.41 $\pm$ 11.46} & \textcolor{gray}{5331.4 $\pm$ 517.54} & \textcolor{gray}{5452.77 $\pm$ 324.07} &         4232.16 $\pm$ 713.9  &      7932.26 $\pm$ 711.87  &         4147.2 $\pm$ 362.87  &   \underline{709.86 $\pm$ 332.07}  &      \textbf{701.61 $\pm$ 199.35} \\
\textsc{Wiki}     & \textcolor{gray}{19692927 $\pm$ 269644} & \textcolor{gray}{18843511 $\pm$ 1553022} & \textcolor{gray}{2634547 $\pm$ 597323} &        2458503 $\pm$ 151593  &      9983566 $\pm$ 434670  &       12410909 $\pm$ 319517  &         \textbf{9532 $\pm$ 10690}  &   \underline{271611 $\pm$ 359320} \\
\bottomrule
\end{tabular}
}
    \end{center}
    \end{table}

\label{sec:num_hyps}

\subsection{Evaluation on financial data } \label{apx:financial_data}

\begin{table}
\begin{center}
\caption{\textbf{Results of neural networks-based methods on the cryptocurrency dataset.} Here, $K = 4$, and the results were averaged over three random seeds. Here, \texttt{TimeMCL} follows the same experimental setup as in the previous benchmark, except that we used Z-Score normalization (instead of mean scaling) during training.}
\label{tab:results_crypto}
\resizebox{\columnwidth}{!}{
\begin{tabular}{l|ccccc||c}
\toprule
{}  & \textcolor{gray}{\textbf{\texttt{Trf.TempFlow}}} & \textcolor{gray}{\textbf{\texttt{Tactis2}}} &  \textbf{\texttt{TimeGrad}}  &   \textbf{\texttt{DeepAR}}  &  \textbf{\texttt{TempFlow}}  &  \textbf{\texttt{TimeMCL(A.)}} \\
\midrule
Distortion       & \textcolor{gray}{\textbf{1334.441 $\pm$ 245.746}} & \textcolor{gray}{1898.896 $\pm$ 255.785} &      1834.202 $\pm$ 188.147  &     2437.798 $\pm$ 94.987  &      1870.915 $\pm$ 174.434  &   \underline{1400.396 $\pm$ 144.5} \\
Total Variation  & \textcolor{gray}{8895.81 $\pm$ 1198.262} & \textcolor{gray}{\underline{4209.638 $\pm$ 1047.498}} &     12797.597 $\pm$ 3040.59  &   32352.73 $\pm$ 3054.369  &     9957.855 $\pm$ 2982.198  &    \textbf{2174.019 $\pm$ 769.041} \\
CRPS             & \textcolor{gray}{\textbf{0.014 $\pm$ 0.001}} & \textcolor{gray}{0.018 $\pm$ 0.001} &           0.018 $\pm$ 0.003  &         0.019 $\pm$ 0.001  &            0.02 $\pm$ 0.001  &      \underline{0.016 $\pm$ 0.004} \\
RMSE             & \textcolor{gray}{\textbf{2275.986 $\pm$ 165.877}} & \textcolor{gray}{\underline{2515.755 $\pm$ 138.424}} &      2642.089 $\pm$ 366.955  &     2743.841 $\pm$ 93.297  &        2756.95 $\pm$ 68.081  &              2528.18 $\pm$ 687.728 \\
FLOPs & \textcolor{gray}{3.89 $\times 10^{7}$} & \textcolor{gray}{9.81 $\times 10^{7}$} &  9.13 $\times 10^{8}$ & \textbf{1.74} $\times \mathbf{10^{5}}$ & 1.94 $\times 10^{7}$ & \underline{9.98 $\times 10^{5}$} \\ 
\bottomrule
\end{tabular}
}
\end{center}
\end{table}

An application of \textbf{\texttt{Time-MCL}} is in forecasting financial time series, such as asset prices. Rather than working directly with raw prices, it is common practice to use the log returns \cite{tsay2005analysis}, defined as $x_t^{d} = \log P_t^{d} - \log P_{t-1}^{d}$, where $P_t^d$ represents the price of asset $d$ at time $t$. These log returns are challenging to forecast. Capturing the extreme tails of their distribution can yield particularly valuable insights for financial applications.

\begin{figure*}[ht]
  \centering
  \begin{minipage}[t]{0.53\textwidth}
    \centering
    \vspace{0pt}%
    \includegraphics[width=0.85\linewidth]{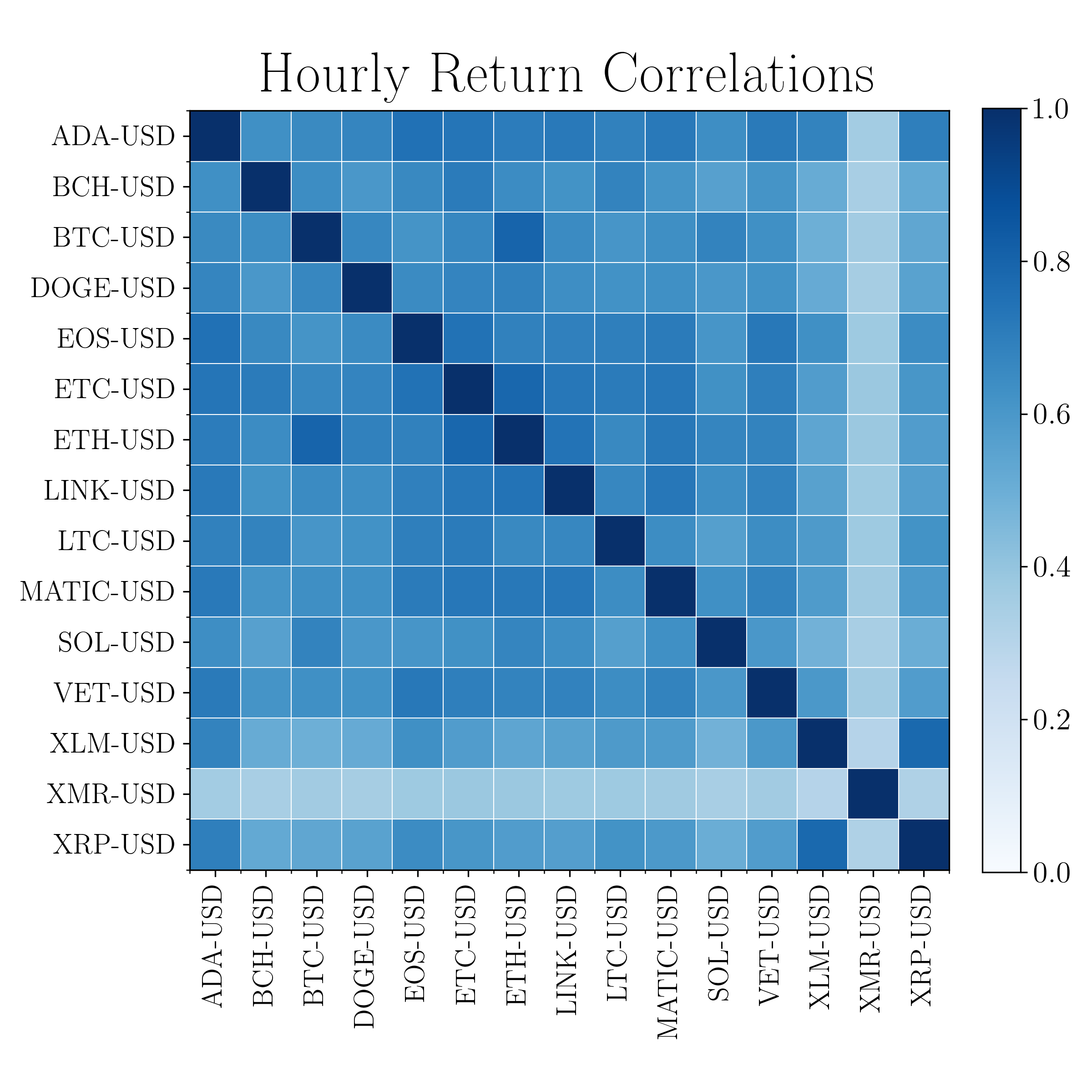}
    \label{fig:corr-matrix}
  \end{minipage}%
  \begin{minipage}[t]{0.44\textwidth}
    \centering
    \vspace{10pt}
    \footnotesize
    \setlength{\tabcolsep}{4pt}
    \label{tab:crypto-summary}
    \begin{tabular}{lrrr}
      \toprule
      \textbf{Ticker} & \textbf{Min.} & \textbf{Max.} & \textbf{Points.}\\
      \midrule
      ADA  &      0.24 &      1.31 & 14\,614\\
      BCH  &    180.45 &    711.92 & 14\,614\\
      BTC  & 25\,001.41 & 108\,240.09 & 14\,614\\
      DOGE &      0.06 &      0.48 & 14\,607\\
      EOS  &      0.40 &      1.50 & 14\,614\\
      ETC  &     14.52 &     39.67 & 14\,614\\
      ETH  &  1\,528.68 &  4\,076.34 & 14\,613\\
      LINK &      5.77 &     30.44 & 14\,614\\
      LTC  &     51.16 &    144.77 & 14\,607\\
      MATIC&      0.26 &      1.28 & 14\,607\\
      SOL  &     17.49 &    287.81 & 14\,613\\
      VET  &      0.01 &      0.08 & 14\,614\\
      XLM  &      0.08 &      0.60 & 14\,614\\
      XMR  &    101.70 &    241.71 & 14\,614\\
      XRP  &      0.40 &      3.39 & 14\,614\\
      \bottomrule
    \end{tabular}
  \end{minipage}

  \caption{\textbf{Cryptocurrency dataset description.} \textit{(Left)} Pair-wise matrix correlations of log returns between the time series.
  \textit{(Right)} Yahoo Finance ticker symbols, number of hourly observations per asset (\textbf{Points}), and the corresponding price scales (\textbf{Min, Max}), thereby summarizing both data volume and cross-asset dependence.}
  \label{fig:crypto_dataset_caracteristics}
\end{figure*}
We obtained cryptocurrency data from \href{https://github.com/ranaroussi/yfinance}{\texttt{YahooFinance}} and was divided into three splits: training (2023-07-01 to 2024-09-01), validation (2024-09-02 to 2024-12-05), and test (2024-12-06 to 2025-03-01), with an hourly resolution. Figure~\ref{fig:crypto_dataset_caracteristics} provides an overview of the assets collected for training, including return correlations among these assets, their price ranges, and the number of data points available. Missing data points were handled using forward filling. 

\textbf{Setup.} We trained and evaluated the previously introduced neural-based baselines on this dataset, using $K = 4$ hypotheses. Each model was trained for $100$ epochs, with $100$ iterations per epoch and a batch size of $64$. The prediction length and context length were both set to $24$. The experimental setup was the same as in the previous benchmark, except that we applied Z-score normalization instead of mean scaling during training, which better handles the wide range of asset price scales. With the exception of \texttt{Tactis2}, which already uses Z-score normalization, we observed that this change significantly improved the performance of all baselines, except for \texttt{DeepAR} where we retained mean scaling. Note that, because the validation losses were monotonically decreasing, we used the final training checkpoint for testing.

\textbf{Results.} Quantitative results on quality (Distortion, CRPS, RMSE), smoothness (Total Variation), and computational cost (FLOPs) are presented in Table~\ref{tab:results_crypto}. \texttt{TimeMCL} produces the smoothest predictions overall. Among non-transformer-based baselines (\texttt{TimeGrad}, \texttt{DeepAR}, and \texttt{TempFlow}), \texttt{TimeMCL} consistently outperforms the others in terms of quality. In this setup, it also performs competitively with \texttt{Tactis2}, which was a strong competitor in the previous benchmark. Unlike in the previous benchmark, we found that the Transformer-based variant of TempFlow (\texttt{Trf.TempFlow}) now outperforms the original \texttt{TempFlow} and slightly surpasses \texttt{TimeMCL} in quality metrics, though it requires nearly $40$ times more FLOPs. In the future, we plan to refine this comparison by implementing a Transformer-based version of \texttt{TimeMCL}.

\begin{figure}
    \centering
    \includegraphics[width=1.0\linewidth]{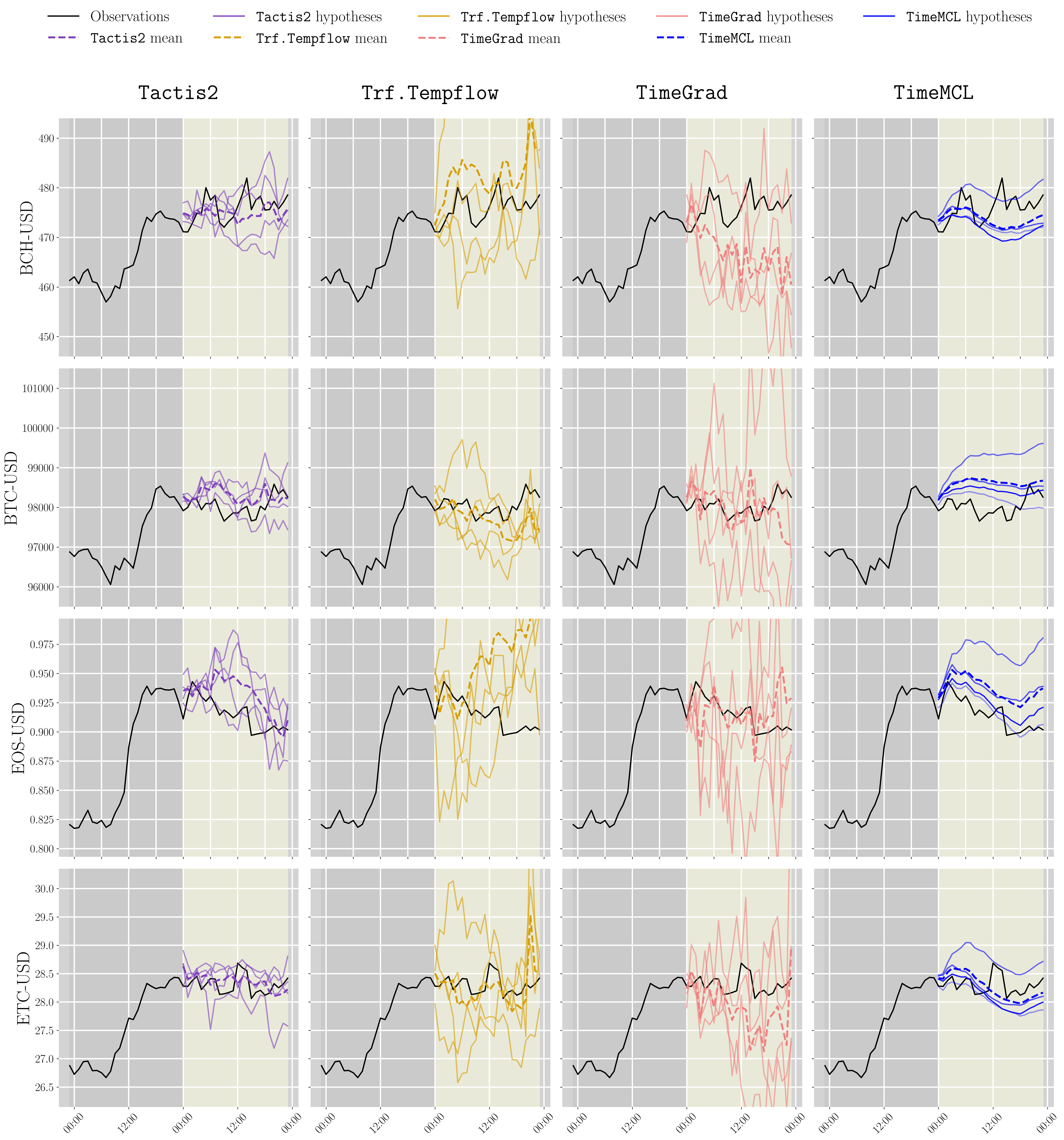}
    \caption{\textbf{Cryptocurrency Price Prediction on 04/01/2025 with $K = 4$ Hypotheses.}}
    \label{fig:crypto_price_prediction}
\end{figure}

Cryptocurrency price predictions are visualized in Figure~\ref{fig:crypto_price_prediction}. Each row represents a target dimension, and each column corresponds to a method listed in Table~\ref{tab:results_crypto}. Following the notation introduced in Figure, \texttt{TimeMCL} predictions are shown in shades of blue, with color intensity reflecting the associated score. We observe that \texttt{TimeMCL} produces smoother predictions compared to other methods and effectively captures different modes in the conditional distribution. For clarity, only a subset of the cryptocurrencies is shown in the figure; however, all models were trained to jointly predict all cryptocurrencies.

\label{secapx:financial_data}

\newpage

\begin{figure}[ht]
    \centering
    \begin{subfigure}
        \centering
        \includegraphics[width=\linewidth]{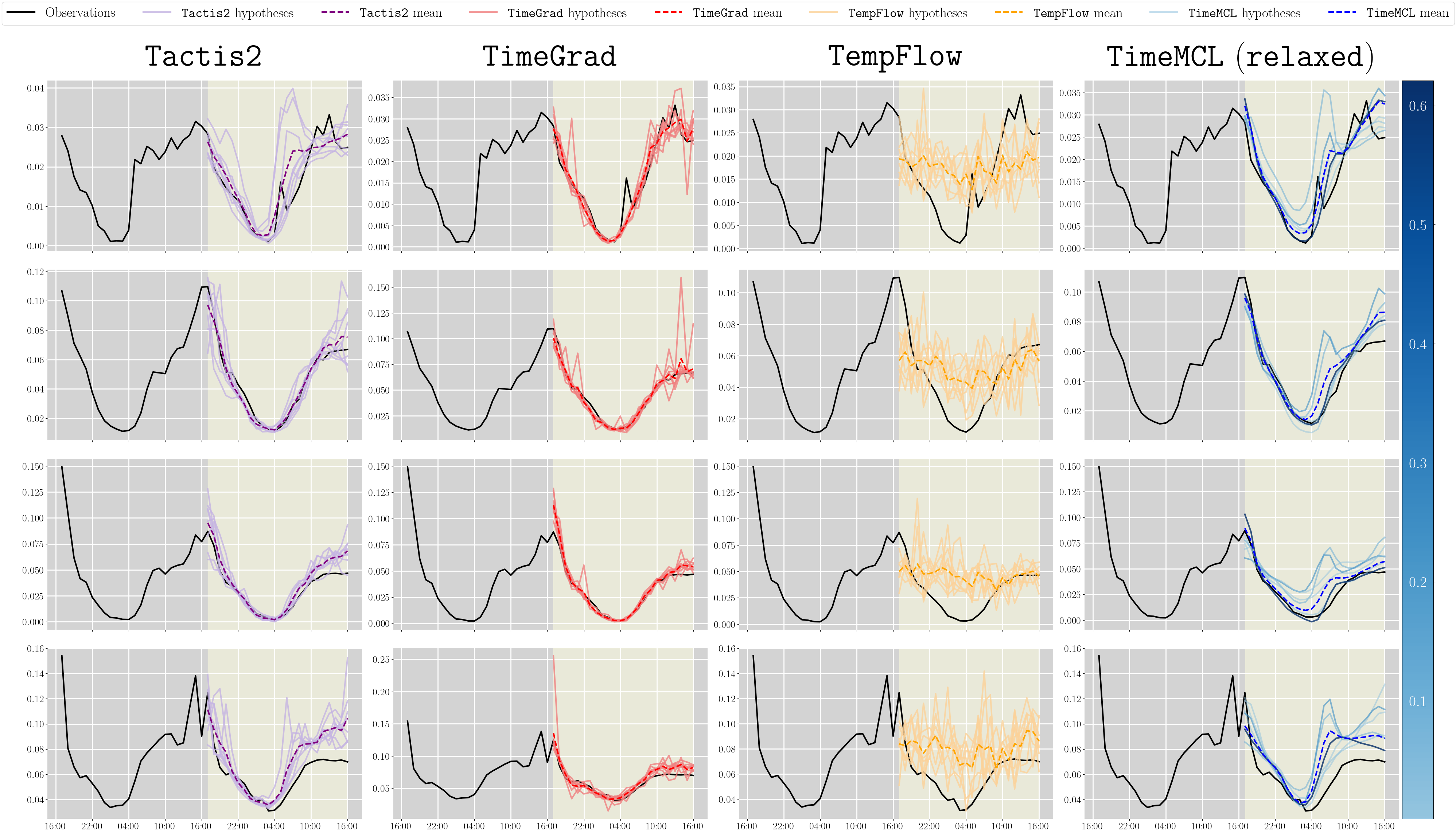}
        \caption{\textbf{Qualitative results on the \textsc{Traffic} dataset}, with same experimental setup as in Figure \ref{fig:combined_plot_main_paper} with $8$ hypotheses.}
        \label{fig:qualitative_traffic}
    \end{subfigure}
    \vspace{1em}
    \begin{subfigure}
        \centering
        \includegraphics[width=\linewidth]{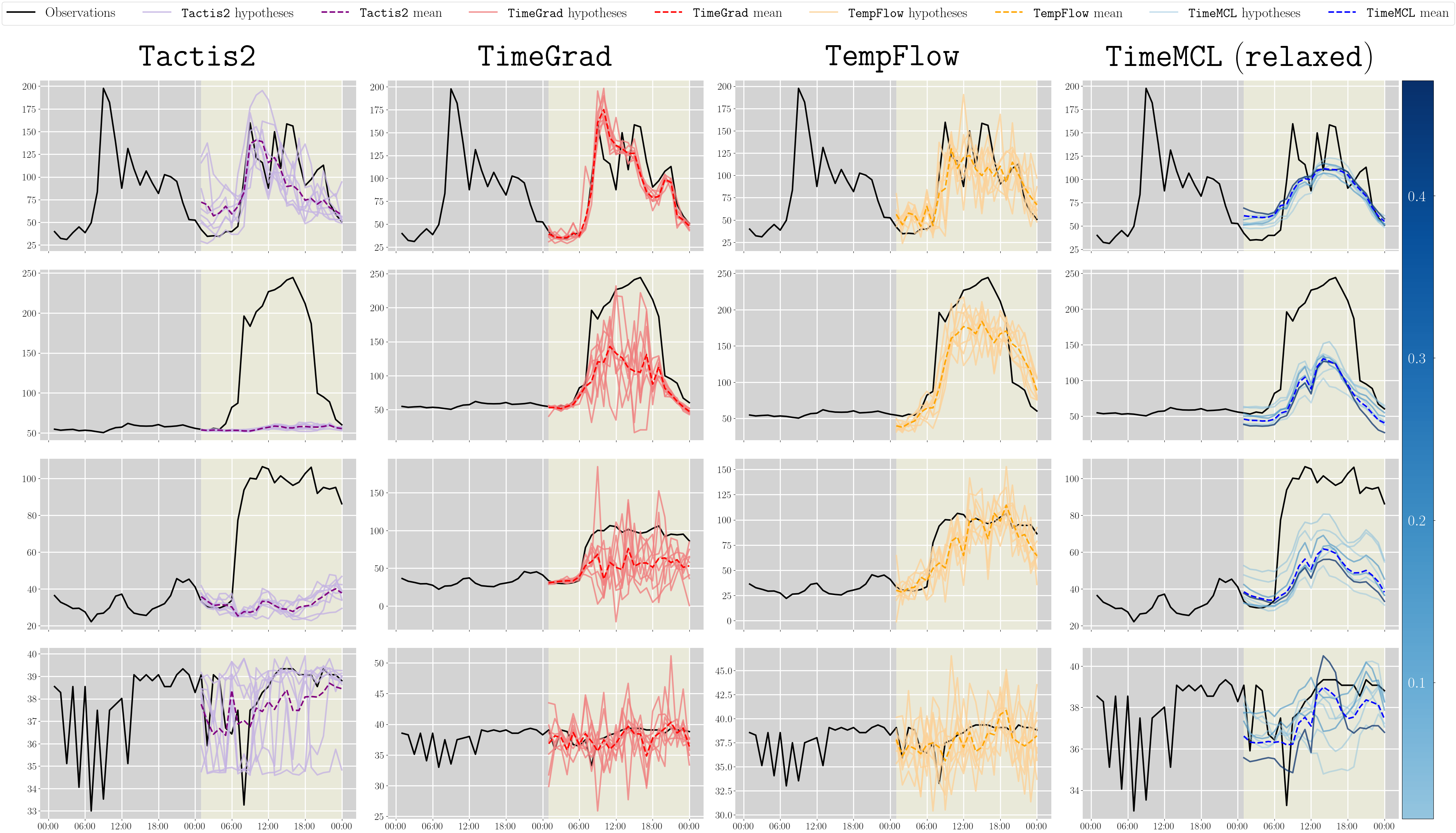}
        \caption{\textbf{Qualitative results on the \textsc{Electricity} dataset}, with same experimental setup as in Figure \ref{fig:combined_plot_main_paper} with $8$ hypotheses.}
        \label{fig:qualitative_elec}
    \end{subfigure}
    \label{fig:qualitative_results_combined}
\end{figure}

\end{document}